\documentclass[letterpaper, 10 pt, conference]{ieeeconf}  

\IEEEoverridecommandlockouts                              

\usepackage{epsfig}
\usepackage{graphicx}
\usepackage{amsmath}
\usepackage{amssymb}
\usepackage{bbm}
\usepackage{booktabs}
\usepackage{multirow}
\usepackage{makecell}
\usepackage{xcolor}
\usepackage{xspace}
\usepackage{colortbl}
\usepackage[colorlinks=black,citecolor=green]{hyperref}

\newcommand{\Name}{Lightning NeRF\xspace}

\title{\LARGE \bf
Lightning NeRF: Efficient Hybrid Scene Representation for Autonomous Driving
}

\author{Junyi Cao$^{1}$, Zhichao Li$^{2}$, Naiyan Wang$^{2}$, and Chao Ma$^{1,\star}$ 
\thanks{$^{1}$ Junyi Cao and Chao Ma are with MoE Key Lab of Artificial Intelligence, AI Institute, Shanghai Jiao Tong University, Shanghai, China.
        {\tt\small \{junyicao, chaoma\}@sjtu.edu.cn}}%
\thanks{$^{2}$ Zhichao Li and Naiyan Wang are with Tusimple, Beijing, China.
        {\tt\small \{leeisabug, winsty\}@gmail.com}}%
\thanks{$^\star$ Chao Ma is the corresponding author.}
}

\makeatletter
\DeclareRobustCommand\onedot{\futurelet\@let@token\@onedot}
\def\@onedot{\ifx\@let@token.\else.\null\fi\xspace}

\def\eg{\emph{e.g}\onedot} 
\def\ie{\emph{i.e}\onedot}

\makeatother

\begin{document}










\def\support{\mbox{support}}
\def\diag{\mbox{diag}}
\def\rank{\mbox{rank}}
\def\grad{\mbox{\text{grad}}}
\def\dist{\mbox{dist}}
\def\sgn{\mbox{sgn}}
\def\tr{\mbox{tr}}
\def\card{{\mbox{Card}}}

\def\balpha{\mbox{{\boldmath $\alpha$}}}
\def\bbeta{\mbox{{\boldmath $\beta$}}}
\def\bzeta{\mbox{{\boldmath $\zeta$}}}
\def\bgamma{\mbox{{\boldmath $\gamma$}}}
\def\bdelta{\mbox{{\boldmath $\delta$}}}
\def\bmu{\mbox{{\boldmath $\mu$}}}
\def\bftau{\mbox{{\boldmath $\tau$}}}
\def\beps{\mbox{{\boldmath $\epsilon$}}}
\def\blambda{\mbox{{\boldmath $\lambda$}}}
\def\bLambda{\mbox{{\boldmath $\Lambda$}}}
\def\bnu{\mbox{{\boldmath $\nu$}}}
\def\bomega{\mbox{{\boldmath $\omega$}}}
\def\bfeta{\mbox{{\boldmath $\eta$}}}
\def\bsigma{\mbox{{\boldmath $\sigma$}}}
\def\bzeta{\mbox{{\boldmath $\zeta$}}}
\def\bphi{\mbox{{\boldmath $\phi$}}}
\def\bxi{\mbox{{\boldmath $\xi$}}}
\def\bvphi{\mbox{{\boldmath $\phi$}}}
\def\bdelta{\mbox{{\boldmath $\delta$}}}
\def\bvarpi{\mbox{{\boldmath $\varpi$}}}
\def\bvarsigma{\mbox{{\boldmath $\varsigma$}}}
\def\bXi{\mbox{{\boldmath $\Xi$}}}
\def\bmW{\mbox{{\boldmath $\mW$}}}
\def\bmY{\mbox{{\boldmath $\mY$}}}

\def\bPi{\mbox{{\boldmath $\Pi$}}}

\def\bOmega{\mbox{{\boldmath $\Omega$}}}
\def\bDelta{\mbox{{\boldmath $\Delta$}}}
\def\bPi{\mbox{{\boldmath $\Pi$}}}
\def\bPsi{\mbox{{\boldmath $\Psi$}}}
\def\bSigma{\mbox{{\boldmath $\Sigma$}}}
\def\bUpsilon{\mbox{{\boldmath $\Upsilon$}}}

\def\mA{{\mathcal A}}
\def\mB{{\mathcal B}}
\def\mC{{\mathcal C}}
\def\mD{{\mathcal D}}
\def\mE{{\mathcal E}}
\def\mF{{\mathcal F}}
\def\mG{{\mathcal G}}
\def\mH{{\mathcal H}}
\def\mI{{\mathcal I}}
\def\mJ{{\mathcal J}}
\def\mK{{\mathcal K}}
\def\mL{{\mathcal L}}
\def\mM{{\mathcal M}}
\def\mN{{\mathcal N}}
\def\mO{{\mathcal O}}
\def\mP{{\mathcal P}}
\def\mQ{{\mathcal Q}}
\def\mR{{\mathcal R}}
\def\mS{{\mathcal S}}
\def\mT{{\mathcal T}}
\def\mU{{\mathcal U}}
\def\mV{{\mathcal V}}
\def\mW{{\mathcal W}}
\def\mX{{\mathcal X}}
\def\mY{{\mathcal Y}}
\def\mZ{{\mathcal{Z}}}


\def\bmA{{\mathbfcal A}}
\def\bmB{{\mathbfcal B}}
\def\bmC{{\mathbfcal C}}
\def\bmD{{\mathbfcal D}}
\def\bmE{{\mathbfcal E}}
\def\bmF{{\mathbfcal F}}
\def\bmG{{\mathbfcal G}}
\def\bmH{{\mathbfcal H}}
\def\bmI{{\mathbfcal I}}
\def\bmJ{{\mathbfcal J}}
\def\bmK{{\mathbfcal K}}
\def\bmL{{\mathbfcal L}}
\def\bmM{{\mathbfcal M}}
\def\bmN{{\mathbfcal N}}
\def\bmO{{\mathbfcal O}}
\def\bmP{{\mathbfcal P}}
\def\bmQ{{\mathbfcal Q}}
\def\bmR{{\mathbfcal R}}
\def\bmS{{\mathbfcal S}}
\def\bmT{{\mathbfcal T}}
\def\bmU{{\mathbfcal U}}
\def\bmV{{\mathbfcal V}}
\def\bmW{{\mathbfcal W}}
\def\bmX{{\mathbfcal X}}
\def\bmY{{\mathbfcal Y}}
\def\bmZ{{\mathbfcal Z}}

\def\0{{\bf 0}}
\def\1{{\bf 1}}

\def\bA{\boldsymbol{A}}
\def\bB{\boldsymbol{B}}
\def\bC{\boldsymbol{C}}
\def\bD{\boldsymbol{D}}
\def\bE{\boldsymbol{E}}
\def\bF{\boldsymbol{F}}
\def\bG{\boldsymbol{G}}
\def\bH{\boldsymbol{H}}
\def\bI{\boldsymbol{I}}
\def\bJ{\boldsymbol{J}}
\def\bK{\boldsymbol{K}}
\def\bL{\boldsymbol{L}}
\def\bM{\boldsymbol{M}}
\def\bN{\boldsymbol{N}}
\def\bO{\boldsymbol{O}}
\def\bP{\boldsymbol{P}}
\def\bQ{\boldsymbol{Q}}
\def\bR{\boldsymbol{R}}
\def\bS{\boldsymbol{S}}
\def\bT{\boldsymbol{T}}
\def\bU{\boldsymbol{U}}
\def\bV{\boldsymbol{V}}
\def\bW{\boldsymbol{W}}
\def\bX{\boldsymbol{X}}
\def\bY{\boldsymbol{Y}}
\def\bZ{\boldsymbol{Z}}

\def\ba{\boldsymbol{a}}
\def\bb{\boldsymbol{b}}
\def\bc{\boldsymbol{c}}
\def\bd{\boldsymbol{d}}
\def\be{\boldsymbol{e}}
\def\bff{\boldsymbol{f}}
\def\bg{\boldsymbol{g}}
\def\bh{\boldsymbol{h}}
\def\bi{\boldsymbol{i}}
\def\bj{\boldsymbol{j}}
\def\bk{\boldsymbol{k}}
\def\bl{\boldsymbol{l}}
\def\bn{\boldsymbol{n}}
\def\bo{\boldsymbol{o}}
\def\bp{\boldsymbol{p}}
\def\bq{\boldsymbol{q}}
\def\br{\boldsymbol{r}}
\def\bs{\boldsymbol{s}}
\def\bt{\boldsymbol{t}}
\def\bu{\boldsymbol{u}}
\def\bv{\boldsymbol{v}}
\def\bw{\boldsymbol{w}}
\def\bx{\boldsymbol{x}}
\def\by{\boldsymbol{y}}
\def\bz{\boldsymbol{z}}

\def\hy{\hat{y}}
\def\hby{\hat{{\bf y}}}

\def\mmE{{\mathbb E}}
\def\mmP{{\mathrm P}}
\def\mmB{{\mathrm B}}
\def\mmR{{\mathbb R}}
\def\mmV{{\mathbb V}}
\def\mmN{{\mathbb N}}
\def\mmZ{{\mathbb Z}}
\def\mMLr{{\mM_{\leq k}}}

\def\tC{\tilde{C}}
\def\tk{\tilde{r}}
\def\tJ{\tilde{J}}
\def\tbx{\tilde{\bx}}
\def\tbK{\tilde{\bK}}
\def\tL{\tilde{L}}
\def\tbPi{\mbox{{\boldmath $\tilde{\Pi}$}}}
\def\tw{{\bf \tilde{w}}}

\def\barx{\bar{\bx}}

\def\pd{{\succ\0}}
\def\psd{{\succeq\0}}
\def\vphi{\varphi}
\def\trsp{{\sf T}}

\def\mRMD{{\mathrm{D}}}
\def \DKL{{D_{KL}}}
\def\st{{\mathrm{s.t.}}}
\def\nth{{\mathrm{th}}}

\def\st{{\mathrm{s.t.}}}
\def\tr{\mathrm{tr}}
\def\grad{{\mathrm{grad}}}

\newtheorem{coll}{Corollary}
\newtheorem{deftn}{Definition}
\newtheorem{thm}{Theorem}
\newtheorem{prop}{Proposition}
\newtheorem{lemma}{Lemma}
\newtheorem{remark}{Remark}
\newtheorem{ass}{Assumption}

\def\red{\textcolor{red}}
\def\blue{\textcolor{blue}}




\maketitle
\thispagestyle{plain}
\pagestyle{plain}

\begin{abstract}
Recent studies have highlighted the promising application of NeRF in autonomous driving contexts. However, the complexity of outdoor environments, combined with the restricted viewpoints in driving scenarios, complicates the task of precisely reconstructing scene geometry. Such challenges often lead to diminished quality in reconstructions and extended durations for both training and rendering. To tackle these challenges, we present \Name. It uses an efficient hybrid scene representation that effectively utilizes the geometry prior from LiDAR in autonomous driving scenarios. \Name significantly improves the novel view synthesis performance of NeRF and reduces computational overheads. Through evaluations on real-world datasets, such as KITTI-360, Argoverse2, and our private dataset, we demonstrate that our approach not only exceeds the current state-of-the-art in novel view synthesis quality but also achieves a five-fold increase in training speed and a ten-fold improvement in rendering speed. Codes are available at \href{https://github.com/VISION-SJTU/Lightning-NeRF}{https://github.com/VISION-SJTU/Lightning-NeRF}.

\end{abstract}

\section{INTRODUCTION}

Neural Radiance Fields (NeRFs)~\cite{mildenhall2021nerf} have paved a novel pathway for novel view synthesis, exhibiting remarkable results across various datasets. When dealing with outdoor scenes, the vastness, complex structures, and fluctuating lighting conditions considerably amplify the complexity of scene reconstruction and substantially heighten the computational demands. Recently, several methods have emerged to address the challenges of applying NeRF to outdoor environments. Specifically, NeRF-W~\cite{martin2021nerf} incorporates a learnable appearance embedding to address lighting variations. In the context of autonomous driving, some techniques~\cite{lu2023urban, guo2023streetsurf, rematas2022urban, yang2023unisim} integrate point clouds to deliver enhanced geometric information, addressing the issue of representing complex structures. 
However, these methods often overlook the efficiency and computational overhead associated with training and rendering. More complex modeling and larger scenes invariably lead to extended model training times. 

In this paper, we propose an efficient hybrid scene representation. We separately model density and color in NeRF, using explicit and implicit approaches respectively. For density, point clouds offer an effective initialization, substantially reducing representational challenges. This allows us to explicitly model density using a limited-resolution voxel grid, eliminating the need for a Multi-Layer Perceptron (MLP). For rendering image details, we retain the implicit modeled color MLP to ensure the capacity to accommodate the highly variable real world. Moreover, we present a more realistic model of the outdoor scene background and color decomposition, further elevating the quality of novel view synthesis and rendering efficiency. Comparative studies on real-world autonomous driving datasets, including KITTI-360~\cite{Liao2022PAMI}, Argoverse2~\cite{wilson2023argoverse} and a private dataset, show that our method not only surpasses current state-of-the-arts in performance for novel view synthesis but also achieves a five-fold improvement in training speed and a ten-fold boost in rendering speed.

\begin{figure}[!t]
    \centering
    \includegraphics[width=.95\linewidth]{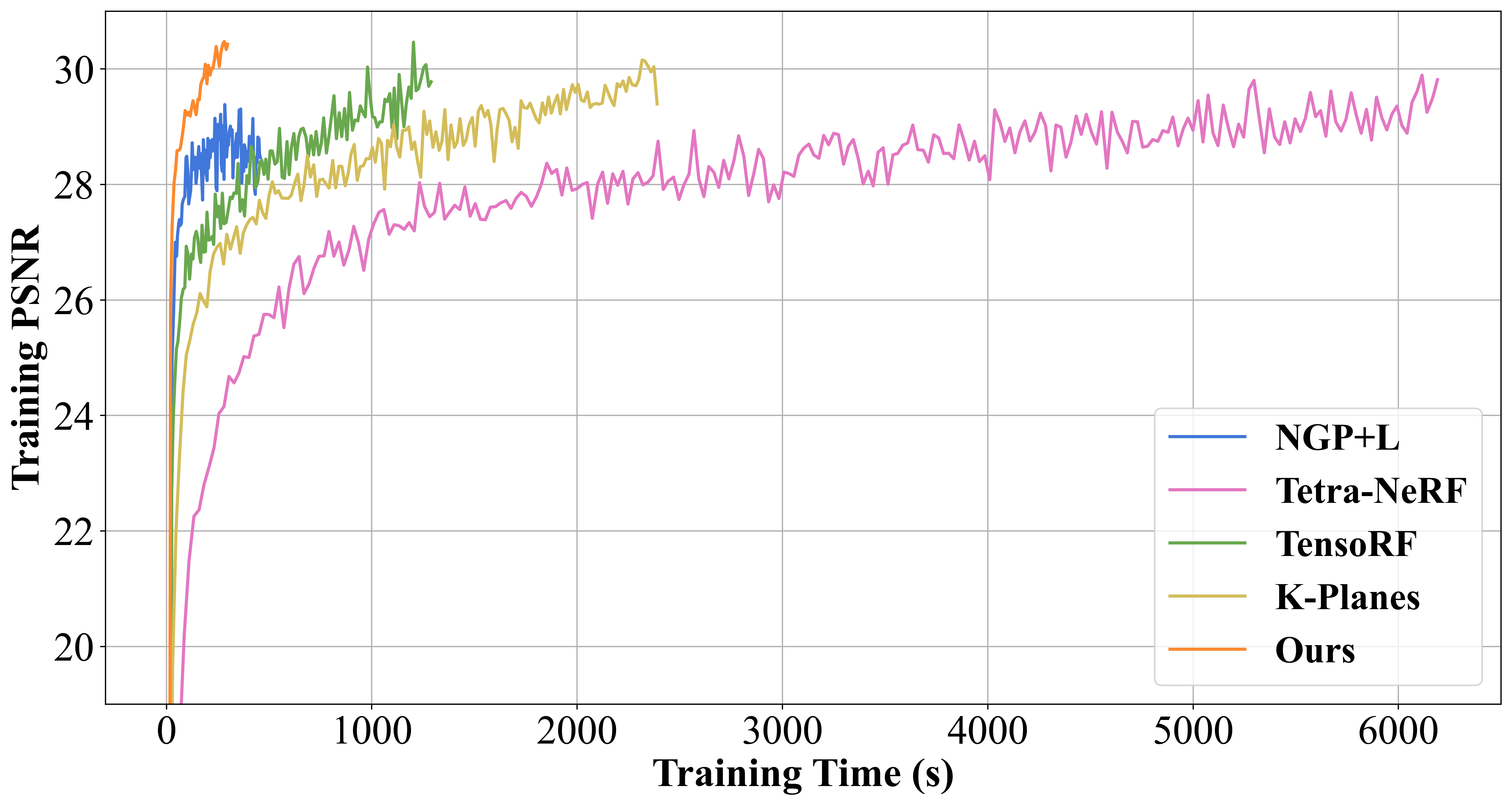}
    \vspace{-3mm}
    \caption{Training efficiency. These curves reflect the trend of training PSNR with time. The values are obtained on sequence 133e2e0b of Argoverse2~\cite{wilson2023argoverse}.}
    \label{fig:cover_img}
\vspace{-4mm}
\end{figure}

\begin{figure*}[!t]
  \centering
  \includegraphics[width=1\linewidth]{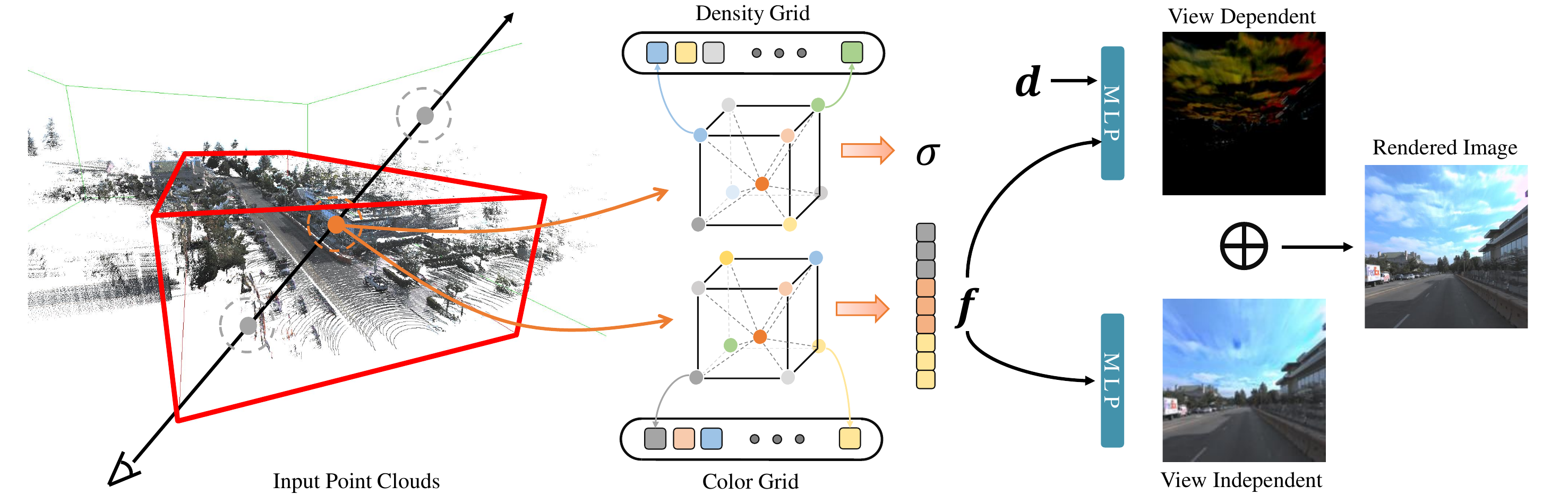}
\vspace{-8mm}
  \caption{Overview of the proposed framework.
  The red and green boxes represent the foreground and background in our proposed scene representation, respectively. 
  Given point cloud data from LiDAR observations, we first use LiDAR Initialization to initialize the scene geometry (see Sec.~\ref{sec:LI}). Then, we query the volume density $\sigma$ and the color embedding feature $\bff$ of each sample point along a ray from the voxel grids (see Sec.~\ref{sec:HSR}). We adopt separate MLPs for modeling view-dependent (with the viewing direction $\bd$ as an additional input) and view-independent colors.
  Combining the two components achieves the final rendered image (see Sec.~\ref{sec:color_decomp}).}
  \label{fig:network}
  \vspace{-3mm}
\end{figure*}

\section{RELATED WORK}

\subsection{Neural Radiance Fields (NeRF)}
NeRF~\cite{mildenhall2021nerf} emerged as a groundbreaking approach to synthesize novel views of a scene by fitting a neural radiance field given calibrated RGB images, offering unprecedented quality in scene representation. Recent works~\cite{zhang2020nerf++,wang2021nerf,lin2021barf,barron2021mip,barron2022mip,pumarola2021d, fu2022panoptic,ye2023featurenerf,chen2023gnesf} pursuit of more detailed modeling, coupled with architectural innovations to push the boundaries of NeRF. However, a significant challenge associated with NeRF is its computational intensity. Specifically, NeRF requires querying a deep MLP millions of times, which results in slow training and rendering. Many studies~\cite{hedman2021baking, muller2022instant} attempt to accelerate the process through more efficient sampling~\cite{liu2020neural, piala2021terminerf, neff2021donerf, xu2022point} or enhanced data structures~\cite{reiser2021kilonerf, garbin2021fastnerf, yu2021plenoctrees, sun2022direct, fridovich2022plenoxels, chen2023mobilenerf, chen2022tensorf, yan2023plenvdb, wang2023fast}, essentially trading space for time. However, when applied to outdoor unbounded scenes, they often encounter problems such as slow convergence, insufficient capacity, or excessive memory overhead.

\subsection{NeRF in Outdoor Scenes}
Outdoor environments, with their expansive scales and complex lighting dynamics, pose fresh challenges for NeRF. To break the limit of single model capacity, Block NeRF~\cite{tancik2022block} strategically segments vast scenes into smaller subsections for more manageable reconstruction. NeRF-W~\cite{martin2021nerf} introduces an appearance embedding for each camera view to learn the photometric differences caused by variations in lighting and viewpoint. To enhance the reconstruction fidelity of large scene geometry, some researches consider additional supervisions, \eg, point cloud data~\cite{lu2023urban,rematas2022urban, yang2023unisim,ost2022neural} or semantic labels~\cite{kundu2022panoptic,zhang2023nerflets}. However, the expansion of scene doesn't merely demand augmented model capacity; it also precipitates a decline in both model convergence and rendering speed. To this end, we harness the geometry cues from point clouds and integrate them with efficient hybrid scene representation. Our method not only facilitates high-fidelity reconstructions, but also effectively boosts the model's convergence and rendering pace.

\section{METHOD}

\subsection{Preliminaries}
NeRF~\cite{mildenhall2021nerf} represents a scene with an implicit function $\mF(\bx, \bd)$, often parameterized by MLPs, that returns a color value $\bc$ and a volume density prediction $\sigma$ for a random 3D point $\bx$ in the scene observed from a viewing direction $\bd$:
\begin{align}
(\sigma, \bff) &= \mM^{\mathrm{(den)}}(\bx),\\
\bc &= \mM^{\mathrm{(rgb)}}(\bff, \bd),
\end{align}
where $\mM^{\mathrm{(den)}},\mM^{\mathrm{(rgb)}}$ are MLPs and $\bff$ is a latent embedding to assist the much shallower $\mM^{\mathrm{(rgb)}}$ to learn $\bc$ (see NeRF++~\cite{zhang2020nerf++} for detailed discussions).  Volume rendering~\cite{max1995optical} is used to synthesize novel views. Concretely, to render a pixel, NeRF leverages hierarchical volume sampling to generate $N$ points along the ray $\br$. The predicted density $\{\sigma_i\}_{i=1}^{N}$ and color feature $\{\bc\}_{i=1}^{N}$ at these positions are combined by accumulation:
\begin{equation}
\label{eqn:volume_render}
\hat{C}(\br) = \sum_{i=1}^N T_i (1-\exp(-\sigma_i\delta_i))\bc_i, T_i = \exp(-\sum_{j=1}^{i-1}\sigma_j\delta_j),
\end{equation}
where $\delta_i$ is the distance between adjacent samples. This process is fully differentiable. Hence, the mean squared error between the predicted and ground-truth colors corresponding to each ray is used to supervise the optimization of NeRF.

While NeRF exhibits superior performance in novel view synthesis, it suffers long training time and slow rendering speed, which is in part because of the inefficiency of its sampling strategy. To this end,~\cite{muller2022instant,liu2020neural,sun2022direct} maintain a coarse occupancy grid during training and only sample locations inside occupied volume. We use a similar sampling strategy as these works to boost the efficiency of our model.

\subsection{Hybrid Scene Representation}
\label{sec:HSR}
Hybrid volumetric representations~\cite{liu2020neural,sun2022direct, chen2022tensorf} have achieved fast optimization and rendering with a compact model. In view of that, we adopt a hybrid voxel-grid representation to model the radiance field for efficiency. 
Briefly, we explicitly model the volume density by storing $\sigma$ at grid vertices while using shallow MLPs to decode color embeddings $\bff$ into final color $\bc$ in an implicit way.
To handle the unbounded nature of outdoor environments, we split our scene representation into two portions for foreground and background separately, as shown in Fig.~\ref{fig:network}. Concretely, we inspect the camera frustum in each frame from a trajectory sequence and define the foreground bounding box so that it tightly wraps all frustums in the aligned coordinate system. 
The background box is obtained by enlarging the foreground box proportionally along each dimension. 

\noindent{\bf Voxel-grid Representation.} A voxel-grid representation explicitly stores the scene properties (\eg, density, RGB color, or feature) in its grid vertices to support efficient feature queries. In this way, for a given 3D location $\bx\in\mmR^3$, we can decode the corresponding property by tri-linear interpolation:
\begin{equation}
\mG(\bx) \triangleq \psi(\bx, \bG): (\mmR^3, \mmR^{C\times N_{\bG} \times N_{\bG} \times N_{\bG}}) \rightarrow \mmR^{C},
\end{equation}
where $\mG(\cdot)$ denotes the decoding operation, $\psi(\cdot)$ is the tri-linear interpolation, and $\bG$ is the voxel grid. $C$ indicates the dimension of the outputs and $N_{\bG}$ gives the grid resolution.

\noindent{\bf Foreground.} We set up two separate feature grids for modeling density and color embedding in the foreground region. Specifically, the density grid mapping $\mG^{(\mathrm{den})}_\mathrm{fg}$  maps positions into density scalars $\sigma$ for volume rendering. For the color-embedding grid mapping $\mG^{(\mathrm{rgb})}_\mathrm{fg}$, we instantiate multiple voxel-grid at different resolution backup by a hash-table, as in~\cite{muller2022instant}, to achieve finer details with affordable memory overhead. The final color embedding $\bff$ is obtained by concatenating outputs over $L$ resolution levels.

\noindent{\bf Background.} Although the aforementioned foreground modeling works well for radiance fields at an object level, extending it to unbounded outdoor scenes is non-trivial. Some related arts like NGP~\cite{muller2022instant} directly expands its scene bounding box so that the background region can be included, while GANcraft~\cite{hao2021gancraft} and URF~\cite{rematas2022urban} introduces a spherical background radiance to handle this issue. However, the former attempt leads to a waste of its capability as most regions inside its scene box are for the background scene.
For the latter proposal, it may fail in handling sophisticated panorama in urban scenes (\eg, undulating buildings or complex landscapes) since it simply assumes the background radiance is only conditioned on view directions.

To this end, we set up an additional background grid model to keep the resolution in our foreground portion unchanged. We adopt the scene parameterization in~\cite{zhang2020nerf++} for the background with deliberate designs. First, unlike its inverse sphere modeling, we use inverse cubic modeling that replaces the $\ell_2$ norm with the $\ell_{\infty}$ norm since we use a voxel-grid representation. Second, we do not instantiate additional MLPs for querying background color to save memory. Specifically, we warp 3D background points into 4D by
\begin{equation}
\bx = (x_1, x_2, x_3) \rightarrow (x_1', x_2', x_3', \frac{1}{r}) = \bx',
\end{equation}
where $r=\|\bx\|_{\infty}$ and $x_k'=x_k/r$ for $k=\{1,2,3\}$.
Then, similar to the grid mapping used in the foreground, we query 4D grid $\mG^{(\mathrm{den})}_\mathrm{bg}$ and $\mG^{(\mathrm{rgb})}_\mathrm{bg}$ given $\bx'$ to obtain $\sigma$ and $\bff$ for background locations. Note that $\mG^{(\mathrm{rgb})}_\mathrm{bg}$ shares the same numbers of levels $L$ and output dimensions $C$ with $\mG^{(\mathrm{rgb})}_\mathrm{fg}$.

In a nutshell, our hybrid scene representation (HSR) outputs density scalars and color embeddings for any 3D locations. The foreground and background decomposition is based on the points' relative positions to the foreground bounding box. Therefore, for simplicity, we will omit the subscript for the grid mapping operation and use $\mG^{(\mathrm{den})}, \mG^{(\mathrm{rgb})}$ in the following sections.
 
\subsection{LiDAR Initialization}
\label{sec:LI}
With our hybrid scene representation, the model can save computation and memory as we query density value directly from the efficient voxel-grid representation rather than the computationally intensive MLP $\mM^{\mathrm{(den)}}$. However, given the large-scale nature and complexity of urban scenes, this lightweight representation easily gets trapped into local minima in optimization due to the limited resolution of the density grid. Fortunately, in autonomous driving scenarios, most Self-Driving Vehicles (SDVs) are equipped with LiDAR sensors, which provide coarse geometry priors for scene reconstruction. To this end, we propose to use the LiDAR point clouds to initialize our density grid to alleviate the handicap of the joint optimization of scene geometry as well as the radiance. Specifically, we first load LiDAR point clouds $P_{\mathrm{pc}}=\{\bp_i|\bp_i\in\mmR^3\}_i^{N_{\mathrm{pc}}}$ and additionally create a set of points $P_\mathrm{bg}=\{\bp_j|\bp_j\in\mmR^3\}_j^{N_{\mathrm{bg}}}$ scattering in the top, front, left and right surfaces of the background box. Then, we manually initialize the density value $\sigma$ at the corresponding location $\bp\in P_\mathrm{pc} \cup P_\mathrm{bg}$ to be a positive constant $\sigma_0$ by adjusting the values in the density grid. Meanwhile, we also update the occupancy grid used for efficient sampling based on the output of the initialized density mapping. This initialization process ensures the sample locations are sparse in the scene and cluster around the objects' surfaces. We find that this step is crucial to the rendering quality and efficiency of our hybrid scene representation (see ablation study in Tab.~\ref{tab:abla}).

 \begin{figure}[!t]
    \centering
    \begin{minipage}{0.48\linewidth}
        \includegraphics[width=\linewidth]{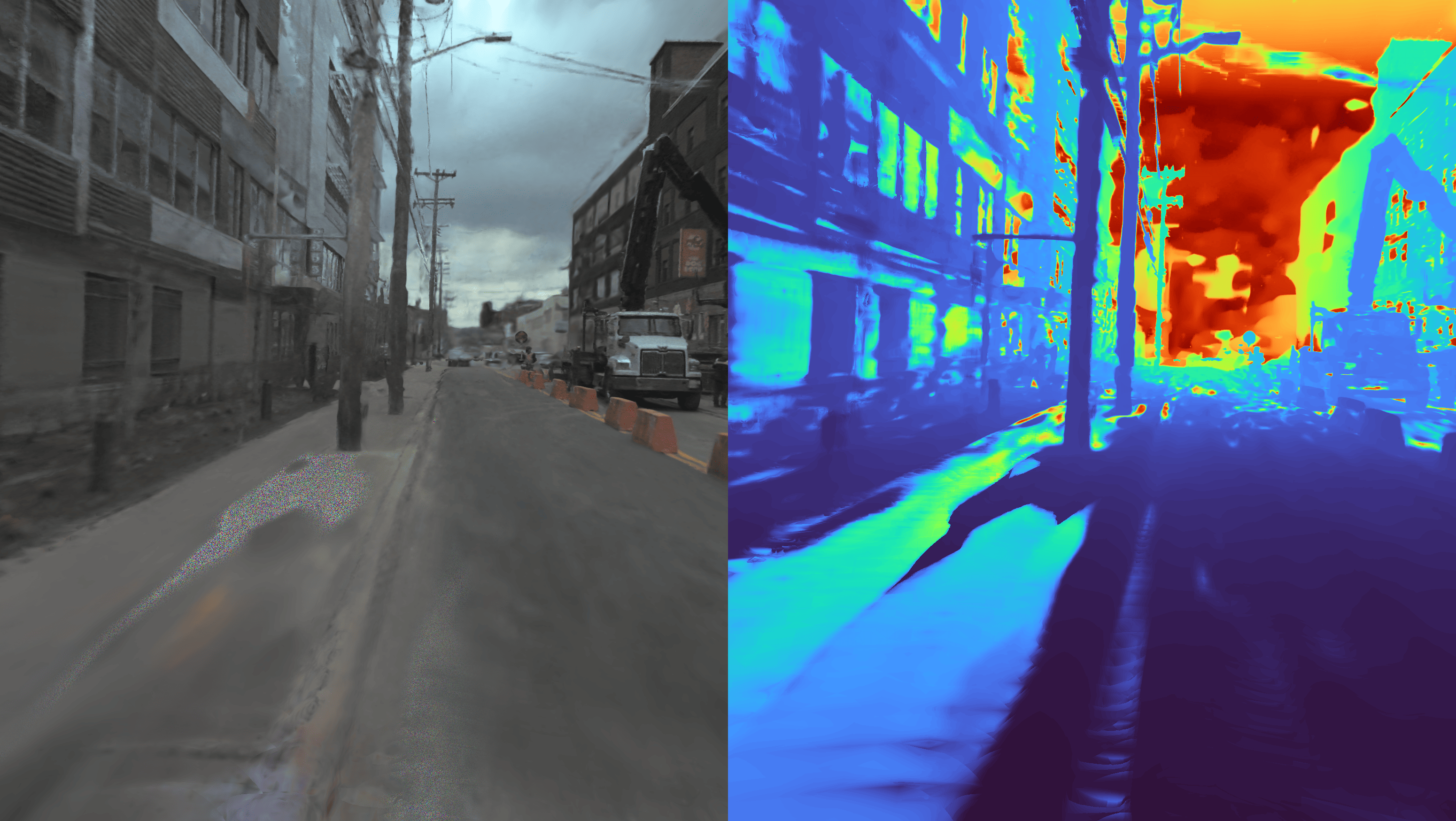}
        \centering w/o CD
    \end{minipage}
    \hfill
    \begin{minipage}{0.48\linewidth}
        \includegraphics[width=\linewidth]{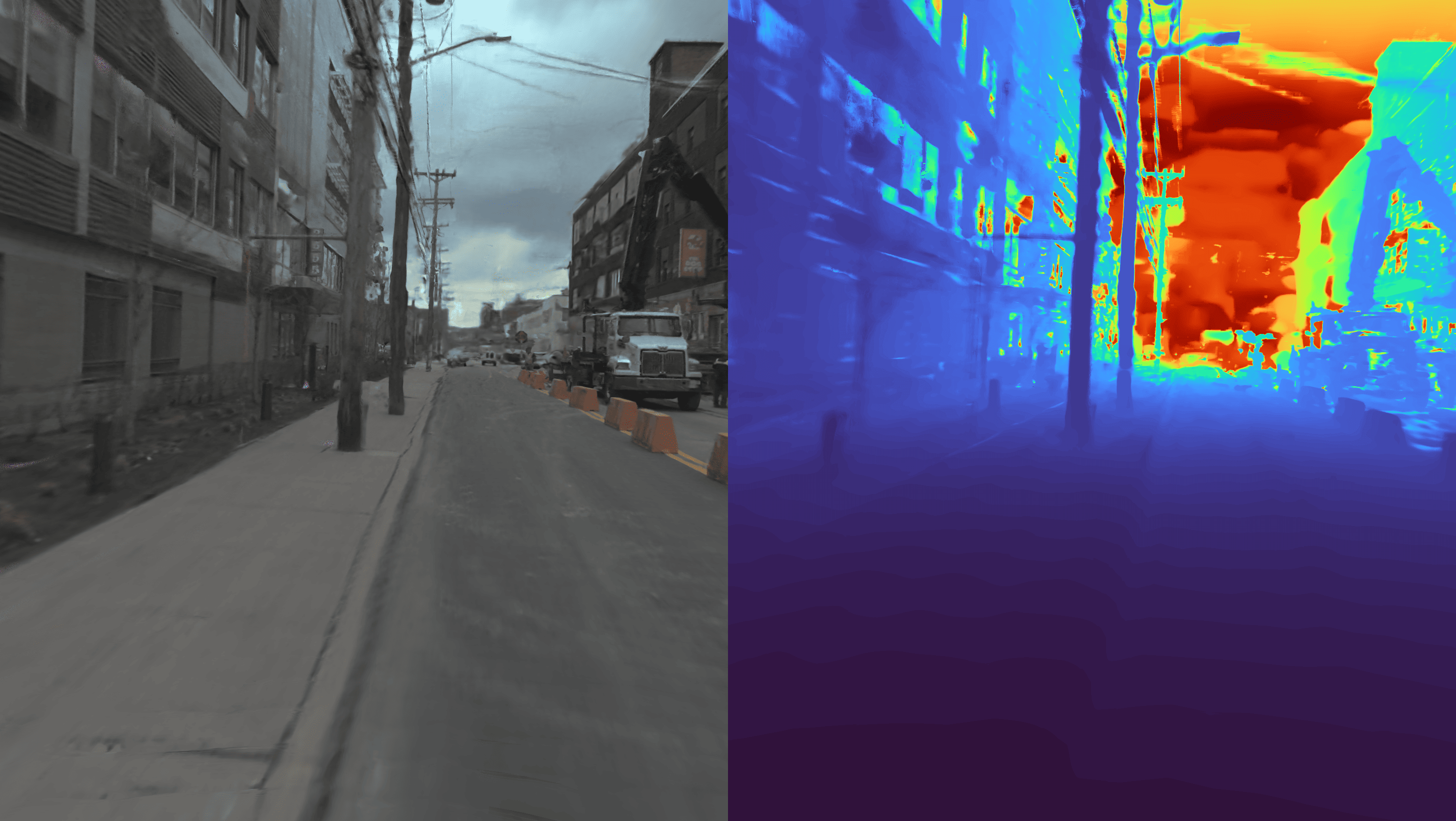}
        \centering w/ CD
    \end{minipage}
    \caption{Visual results in the extrapolation setting with or without the proposed color decomposition (CD). Best viewed in color with zoomed-in.}
    \label{fig:view_cd}
\vspace{-5mm}
\end{figure}

\subsection{Color Decomposition}
\label{sec:color_decomp}

The original NeRF~\cite{mildenhall2021nerf} uses a view-dependent MLP $\mM^{\mathrm{(rgb)}}$ to model the color in the radiance field, which is a simplification of the physical world where radiance is composed of diffuse (view-independent) color and specular (view-dependent) color~\cite{phong1998illumination,greenberg1997framework,zhang2021nerfactor,pharr2023physically}. Besides, since the final output color $\bc$ is fully entangled with the viewing direction $\bd$, it poses difficulties to render high-fidelity images in unseen views (\ie, extrapolation), as $\mM^{\mathrm{(rgb)}}$ is not optimized for viewing directions out of training domains. However, this is the common case in autonomous driving since single trip usually can only scan limited space. Thus, we propose to decompose the color into view-dependent factor $\bc_\mathrm{vd}$ and view-independent factor $\bc_\mathrm{vi}$ to better simulate the physical world. In this way, when the model faces large shifts in viewing directions, the view-independent factor can still give a reasonable prediction for the final color. As shown in Fig.~\ref{fig:view_cd}, our method trained without color decomposition (CD) fails in novel view synthesis in the extrapolation setting (\ie, shift the viewing direction 2 meters left based on training views)
while ours with color decomposition gives a plausible rendering result.

We introduce two lightweight MLPs, denoted as $\mM^{(\mathrm{rgb\_vd})}, \mM^{(\mathrm{rgb\_vi})}$, for modeling view-dependent color $\bc_\mathrm{vd}$ and view-independent color $\bc_\mathrm{vi}$, respectively:
\begin{align}
\bc_\mathrm{vd} &= \mM^{\mathrm{(rgb\_vd)}}(\bff, \bd), \\
\bc_\mathrm{vi} & = \mM^{(\mathrm{rgb\_vi})}(\bff).
\end{align}
The final color at the sample location is the sum of these two factors:
\begin{equation}
    \bc = \bc_\mathrm{vd} + \bc_\mathrm{vi}.
\end{equation}
We then use the conventional volume rendering to obtain the color for a pixel as in Eqn. (\ref{eqn:volume_render}).

\subsection{Training Loss}
We modify the photometric loss with a re-scaling weight $w_i$ to optimize our model to focus on the hard samples for fast convergence. The weight coefficient is defined as:
\begin{equation}
w_i = \mathrm{clamp}(\tilde{w}_i),
\end{equation}
where $\mathrm{clamp}(\cdot)$ means to clip the value into $[1,10]$ and
\begin{equation}
\tilde{w}_i = \frac{\|\hat{C}(\br_i) - C(\br_i)\|_2^2}{\min_{i\in \{1,...,N\}}(\|\hat{C}(\br_i) - C(\br_i)\|_2^2)}.
\end{equation}
Our weighted photometric loss is written as:
\begin{equation}
\mL_\mathrm{p} = \frac{1}{N} \sum_{i=1}^N \langle w_i \rangle \cdot \|\hat{C}(\br_i) - C(\br_i)\|_2^2,
\end{equation}
where $\hat{C}(\br), C(\br)$ are the predicted and the ground-truth color, $\langle \cdot \rangle$ denotes the stop-gradient~\cite{chen2021exploring} operation. We additionally regularize the view-dependent color $\bc_\mathrm{vd}$ with a $\ell_1$ regularizer during training:
\begin{equation}
    \mL_\mathrm{r} = \frac{1}{M} \sum_{j=1}^{M} \|\bc_{\mathrm{vd}, j}\|_1.
\end{equation}
The total loss function is:
\begin{equation}
\label{eqn:total_loss}
    \mL= \mL_\mathrm{p} + \lambda \mL_\mathrm{r},
\end{equation}
where $\lambda$ is a weight parameter for balancing the two losses.

\begin{figure*}[!t]
    \centering
    \begin{minipage}{0.19\textwidth}
        \includegraphics[width=\linewidth]{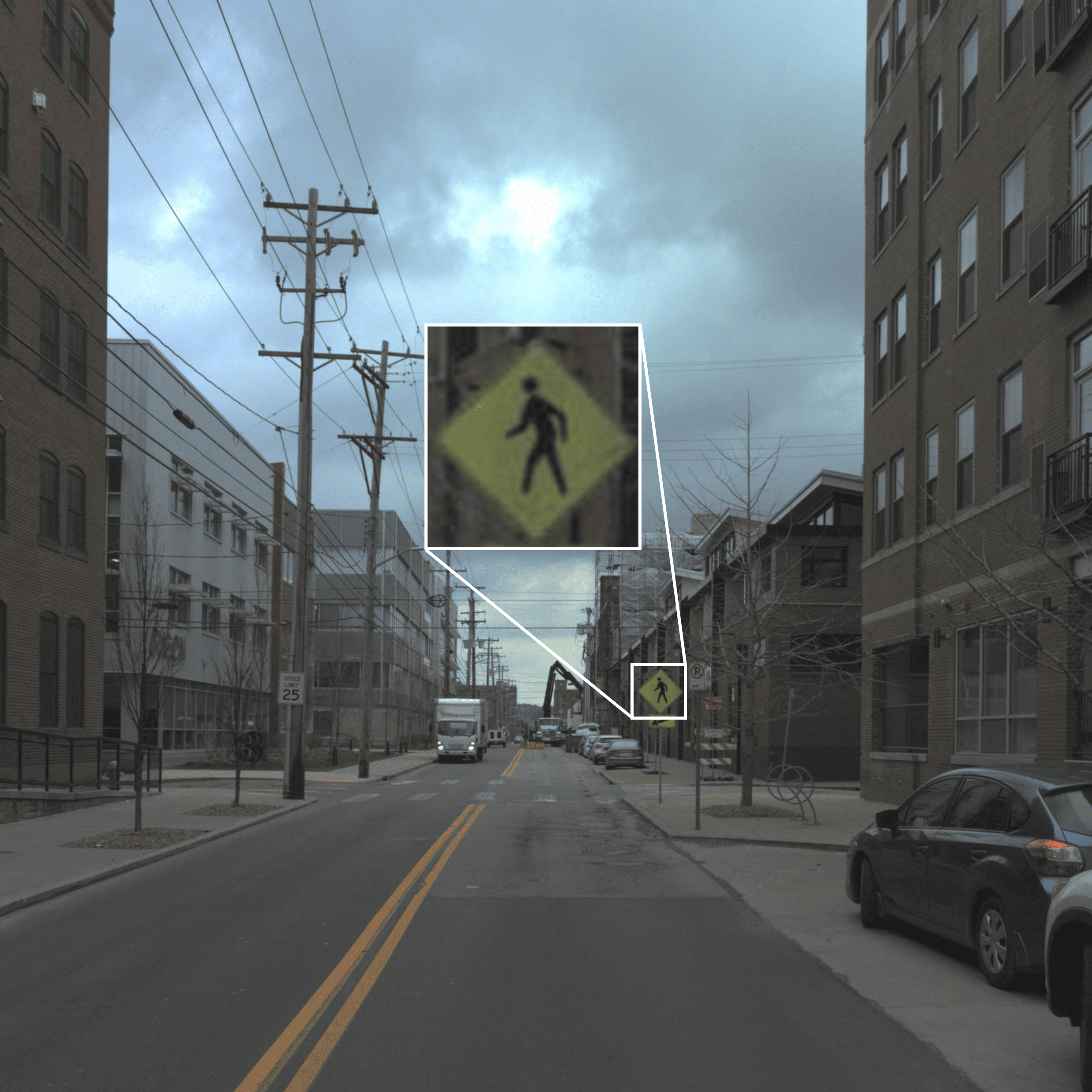}
    \end{minipage}
    \begin{minipage}{0.19\textwidth}
        \includegraphics[width=\linewidth]{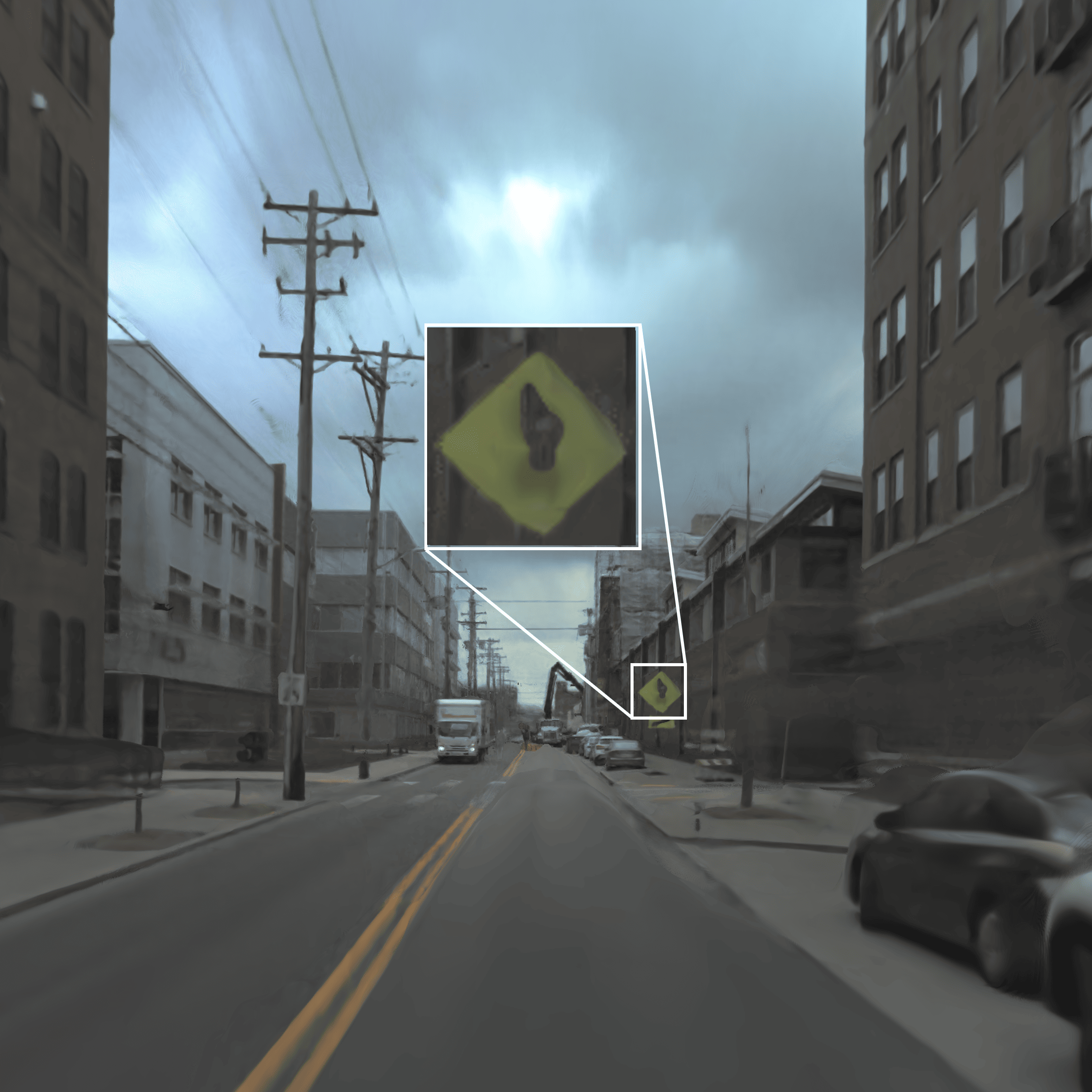}
    \end{minipage}
    \begin{minipage}{0.19\textwidth}
        \includegraphics[width=\linewidth]{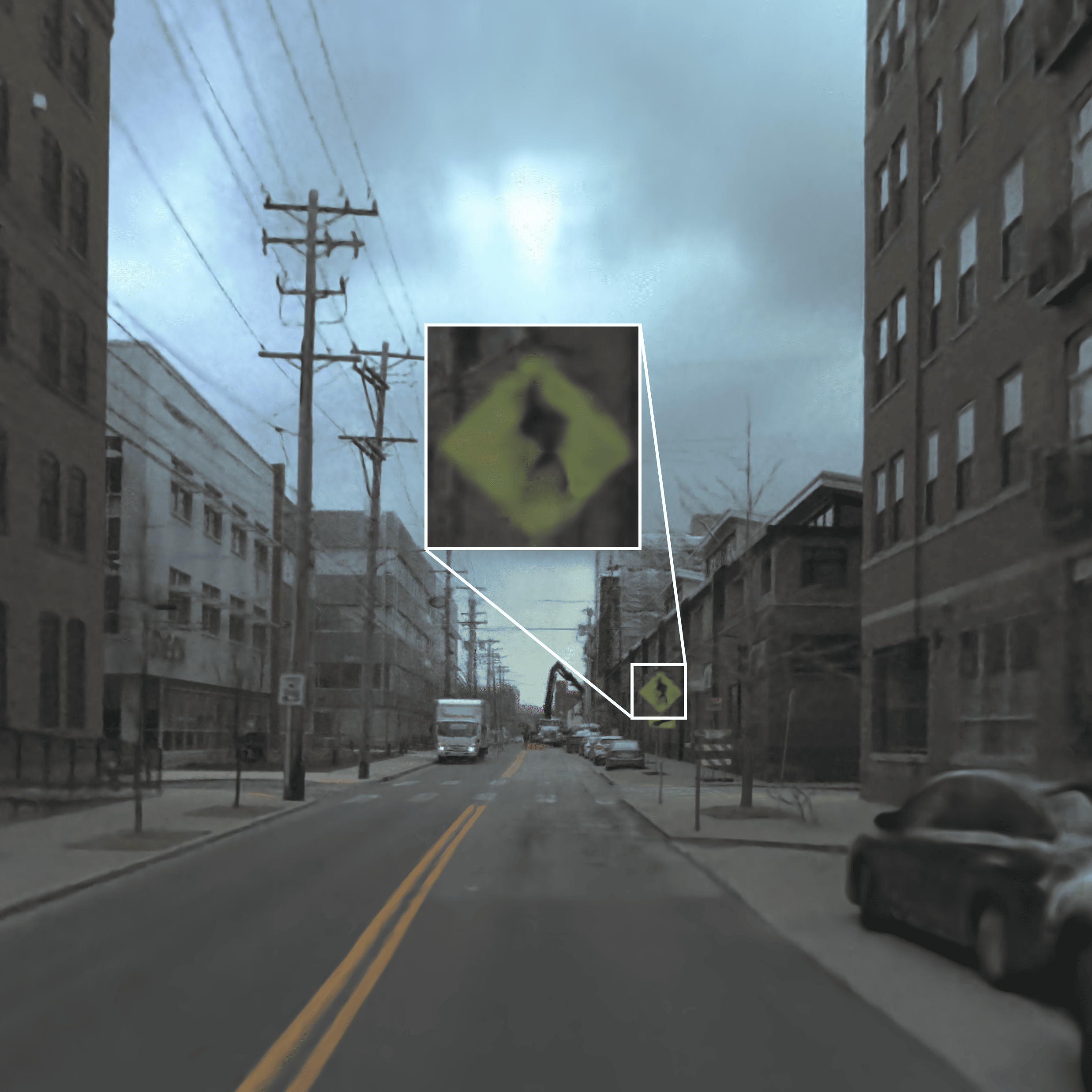}
    \end{minipage}
    \begin{minipage}{0.19\textwidth}
        \includegraphics[width=\linewidth]{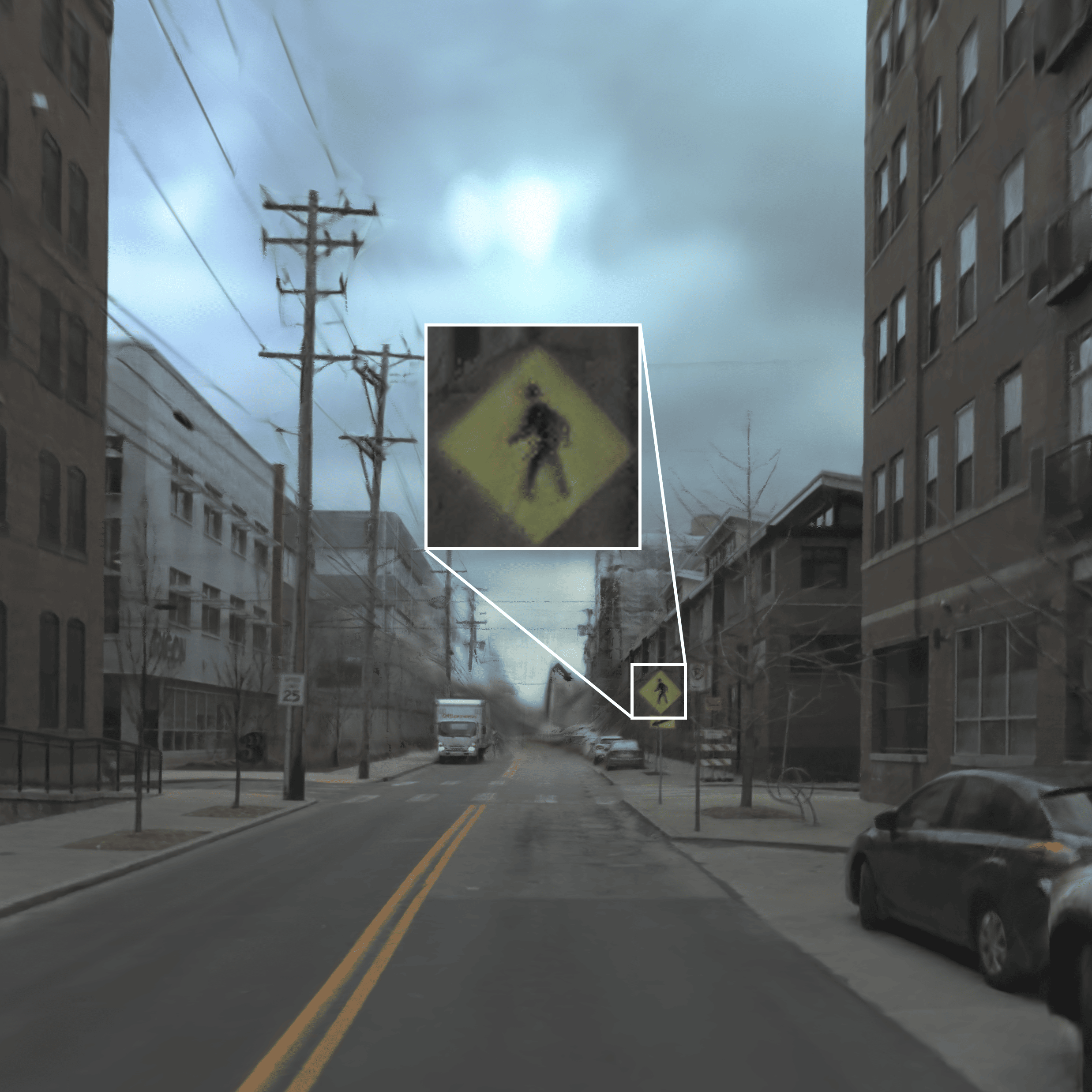}
    \end{minipage}
    \begin{minipage}{0.19\textwidth}
        \includegraphics[width=\linewidth]{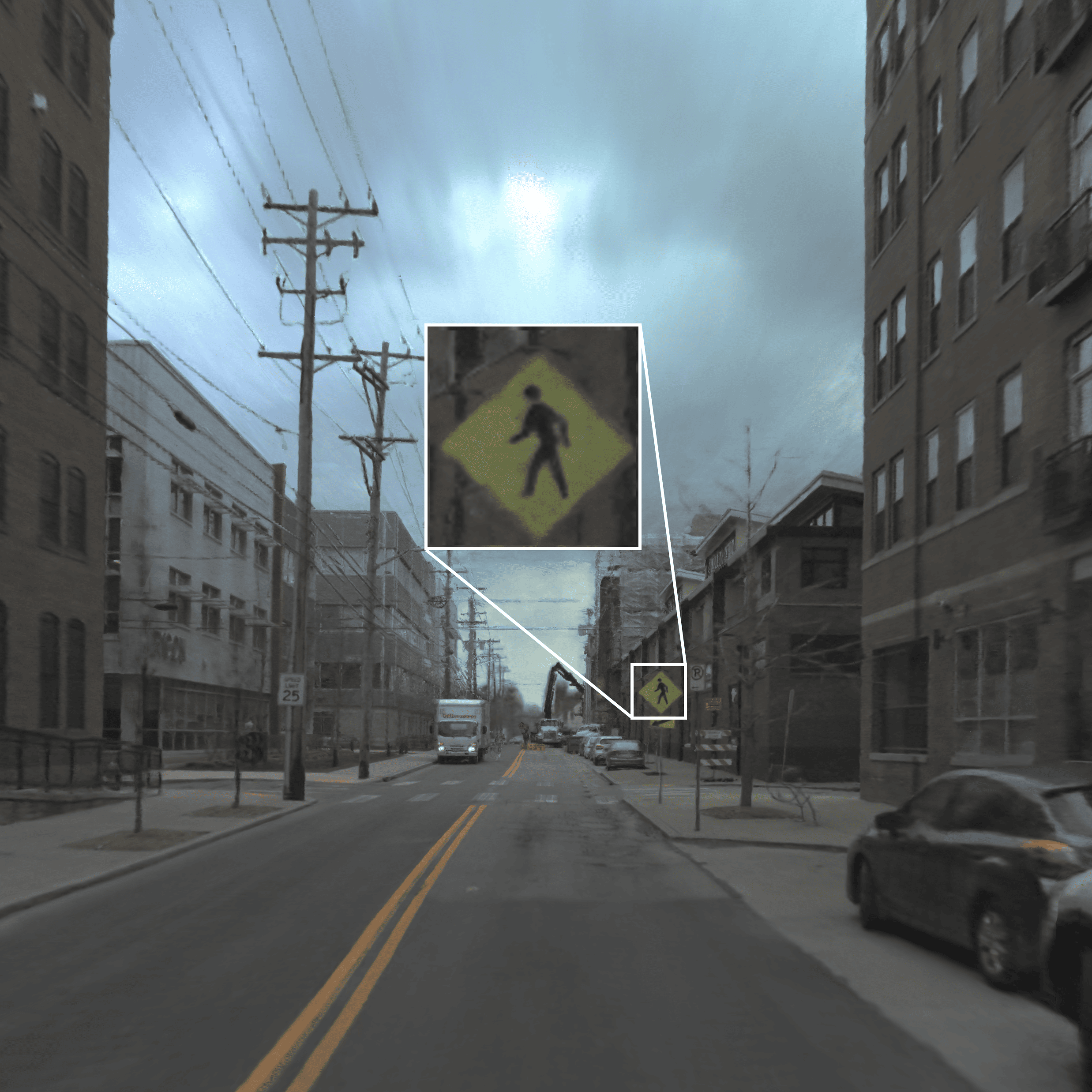}
    \end{minipage}

    \vspace{2mm}
    \begin{minipage}{0.19\textwidth}
        \includegraphics[width=\linewidth]{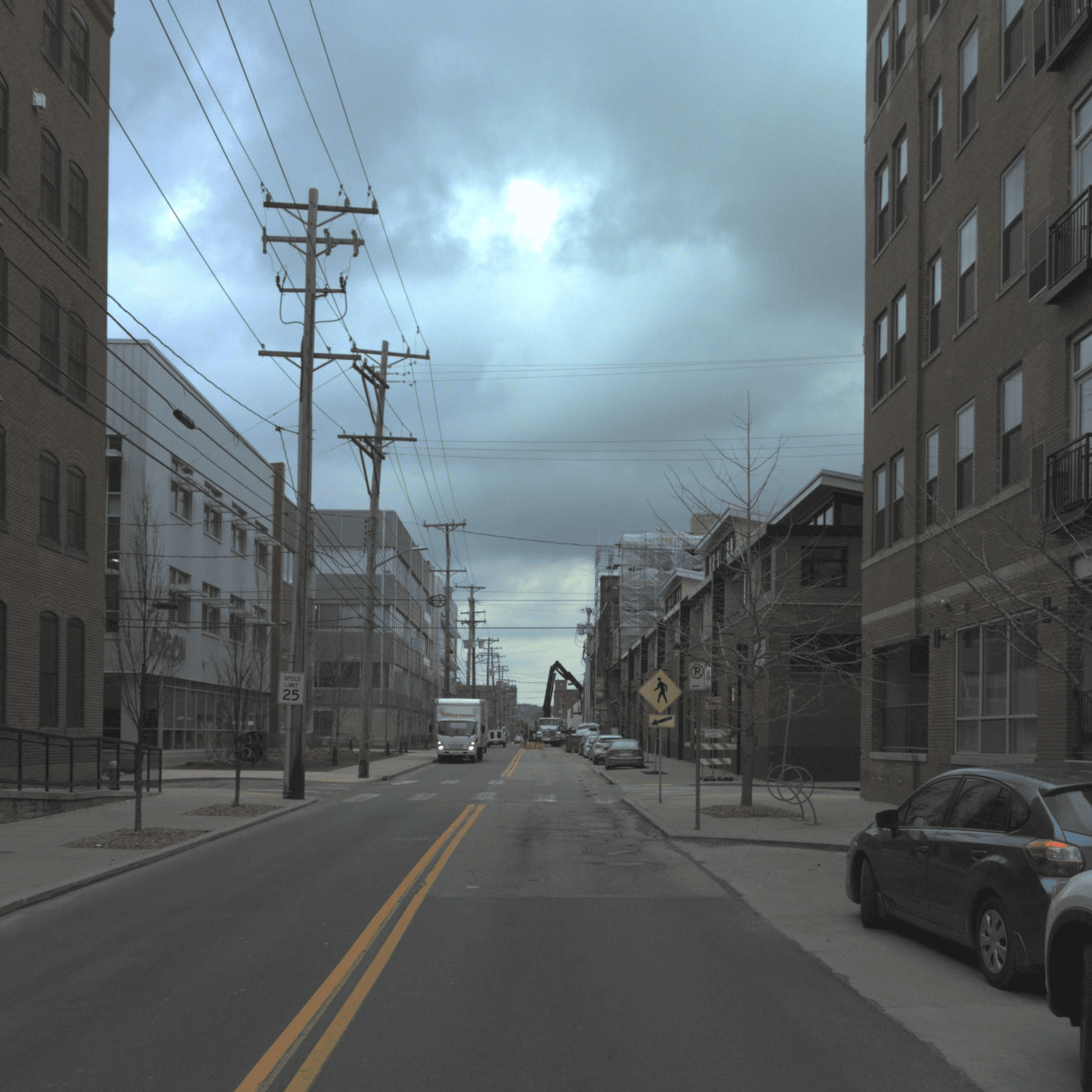}
    \end{minipage}
    \begin{minipage}{0.19\textwidth}
        \includegraphics[width=\linewidth]{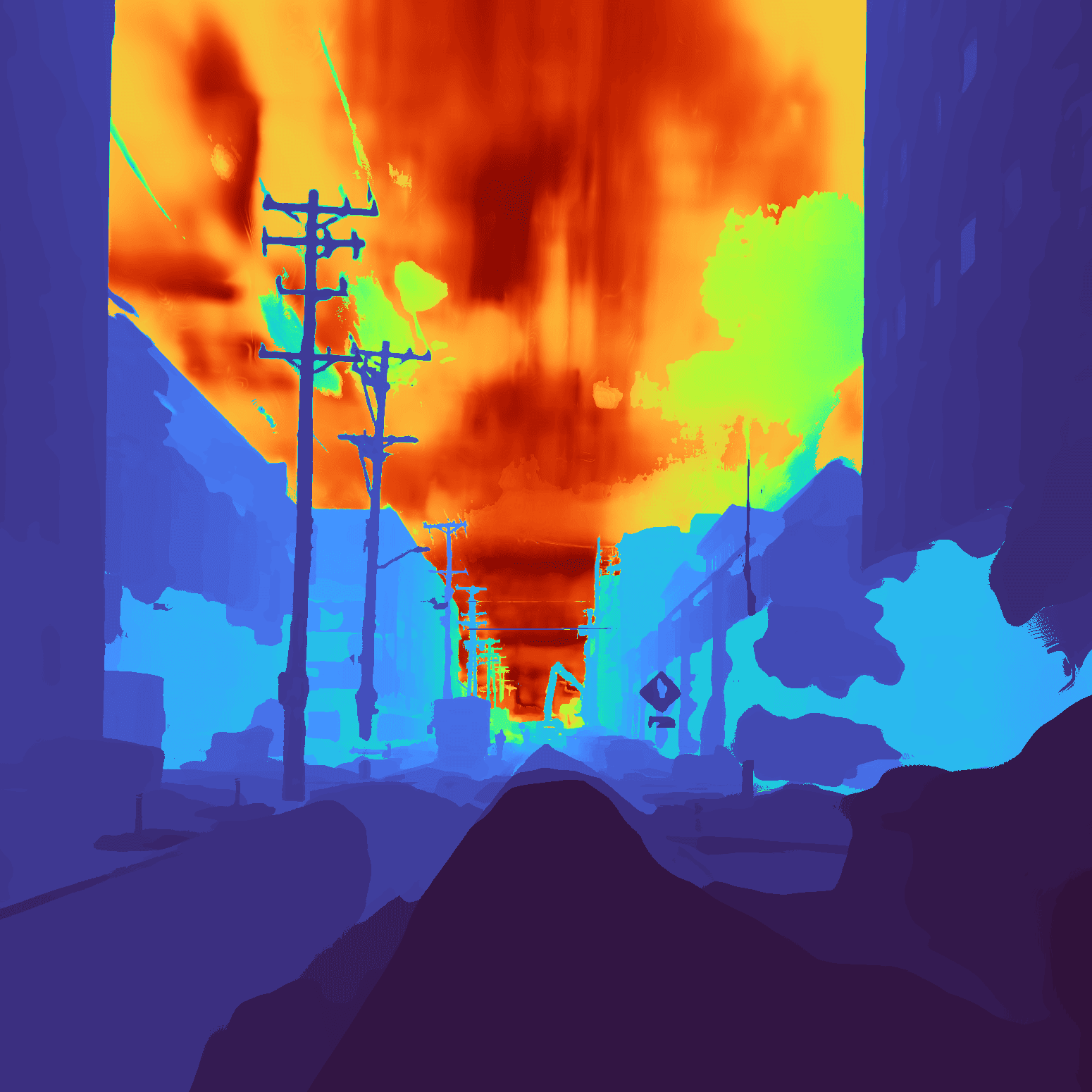}
    \end{minipage}
    \begin{minipage}{0.19\textwidth}
        \includegraphics[width=\linewidth]{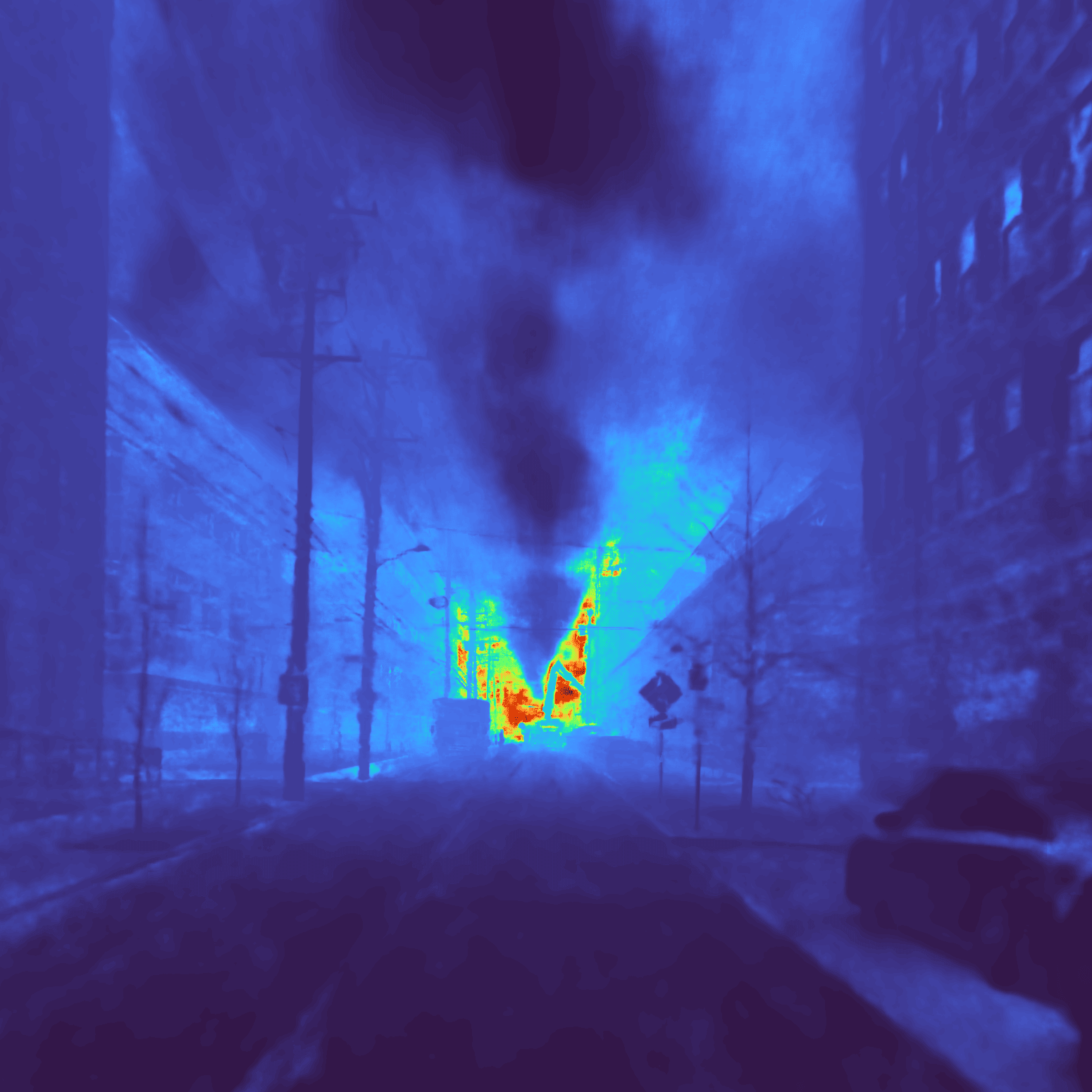}
    \end{minipage}
    \begin{minipage}{0.19\textwidth}
        \includegraphics[width=\linewidth]{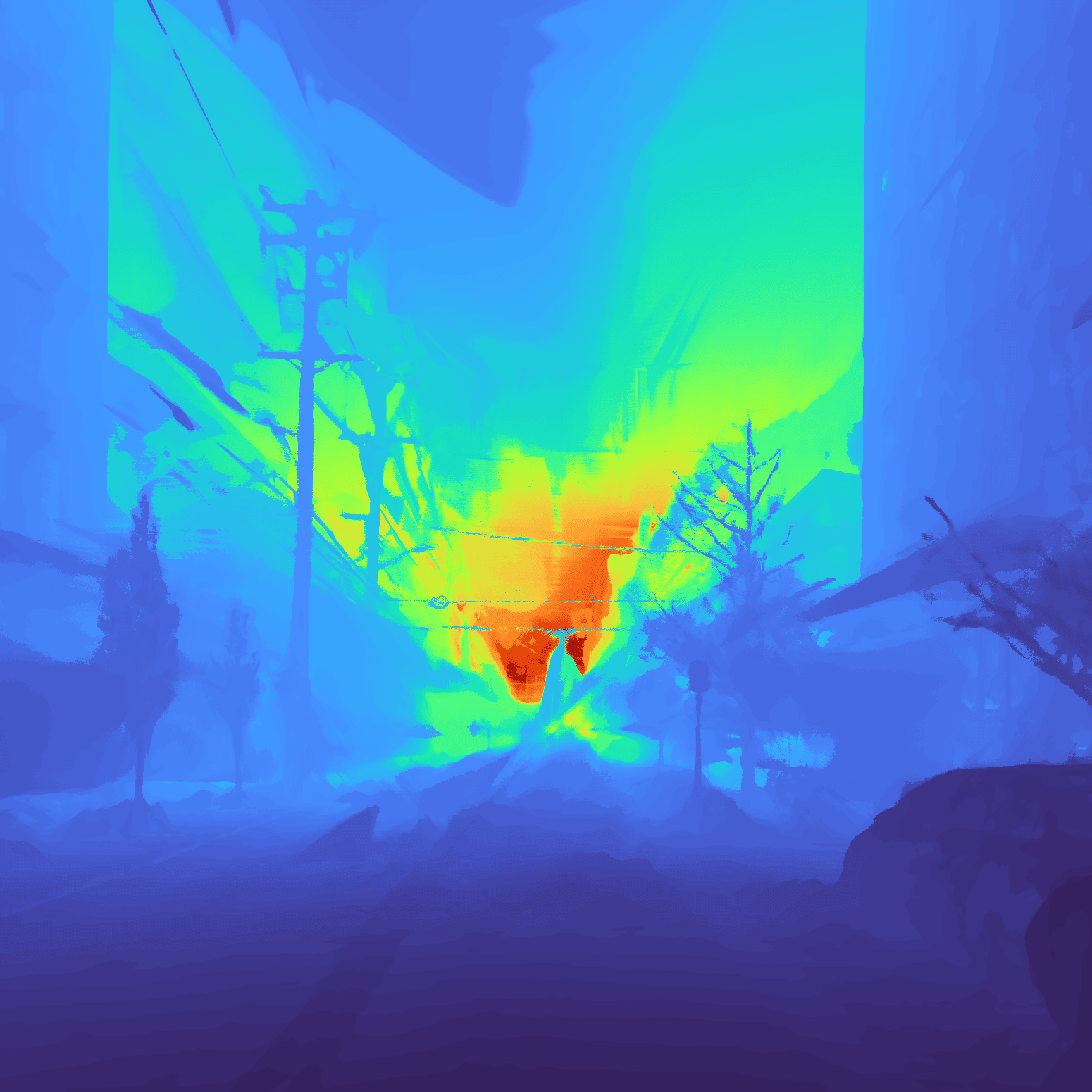}
    \end{minipage}
    \begin{minipage}{0.19\textwidth}
        \includegraphics[width=\linewidth]{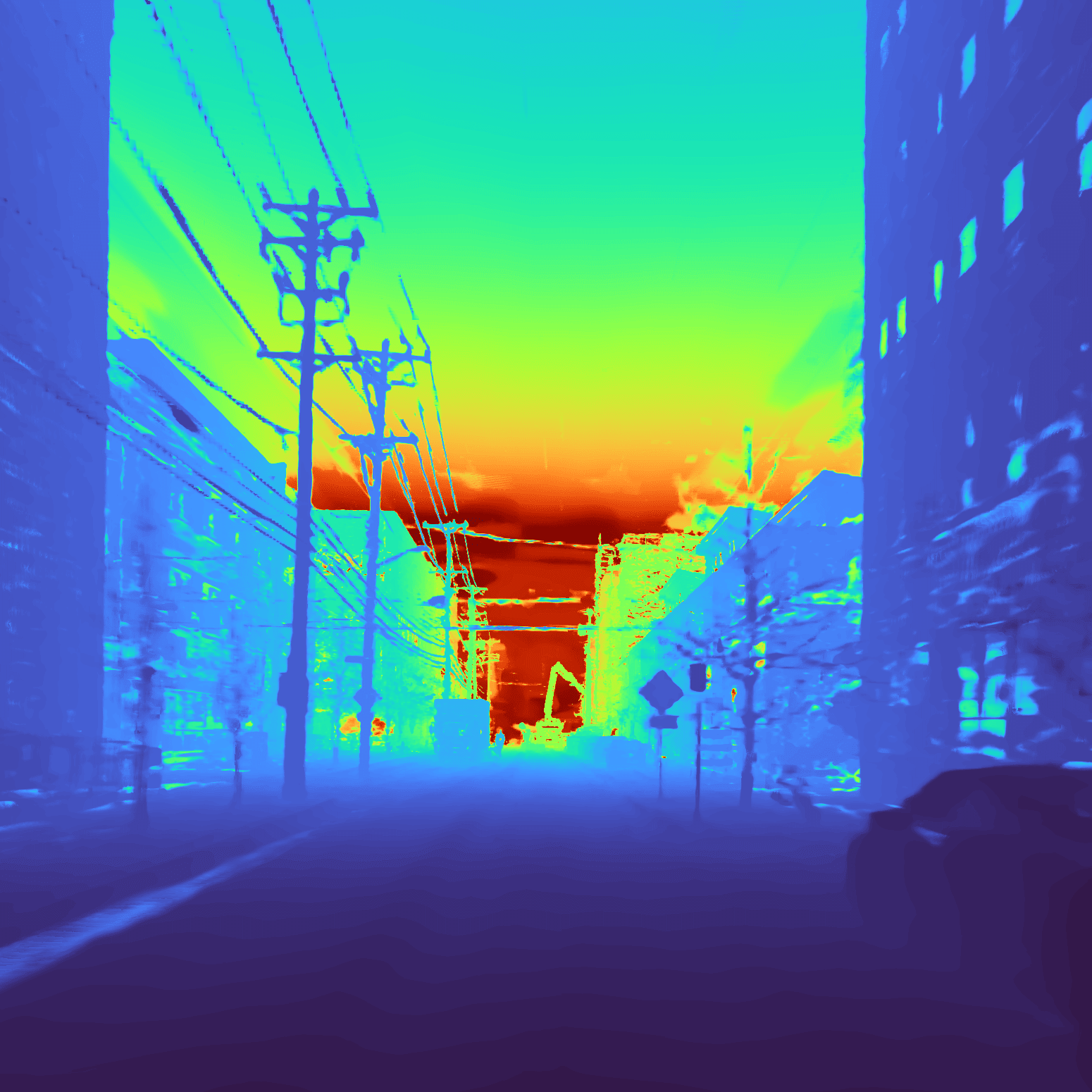}
    \end{minipage}
    
    \vspace{1mm}

    \begin{minipage}{0.19\linewidth}
        \centering
        Ground Truth
    \end{minipage}
    \begin{minipage}{0.19\linewidth}
        \centering
        K-Planes~\cite{fridovich2023k}
    \end{minipage}
    \begin{minipage}{0.19\linewidth}
        \centering
        NGP~\cite{muller2022instant}+L~\cite{rematas2022urban}
    \end{minipage}
    \begin{minipage}{0.19\linewidth}
        \centering
        Tetra-NeRF~\cite{kulhanek2023tetra}
    \end{minipage}
    \begin{minipage}{0.19\linewidth}
        \centering
        Ours
    \end{minipage}
    \vspace{-2mm}
    \caption{Qualitative results on the Argoverse2. The first row shows rendered images and the second row shows depth maps. Best viewed in color.}
    \label{fig:dep}
\end{figure*}

\section{EXPERIMENTS}

\subsection{Experimental Setup}
\noindent{\bf Datasets.}
Three urban datasets are considered in our experiments. Two of them are public-available, \ie, KITTI-360~\cite{Liao2022PAMI} and Argoverse2~\cite{wilson2023argoverse}, while the remaining one is a private dataset used for evaluation on long trajectories. For KITTI-360, we follow~\cite{lu2023urban} to select 5 scenes mainly containing static objects. Note that LiDAR sensors used in KITTI-360 only capture near-ground objects due to a limited vertical field of view. 
The incomplete LiDAR data inevitably hampers the optimization of our model. 
To test our method with complete LiDAR data, we use Argoverse2 and the private dataset. We select 6 sequences from Argoverse2 \emph{Sensor Dataset} and 2 trajectories from the private dataset. We select every 10th image in the sequences as the test data for these two datasets and take the rest as the training data.

\noindent{\bf Implementation Details.} Our model is based on the NeRFStudio framework~\cite{tancik2023nerfstudio}. During training, we use RAdam~\cite{liu2019variance} optimizer for the density and color embedding grids with an initial learning rate of 1.0. The MLP networks are optimized via Adam~\cite{kingma2014adam} optimizer with an initial learning rate of 0.01.
$\lambda$ in Eqn. (\ref{eqn:total_loss}) is set to be 0.01. 
We use a batch size of 65,536 rays for training 30k iterations unless otherwise specified. All experiments are conducted on an NVIDIA RTX 3090 GPU.

\noindent{\bf Evaluation Metrics.} Following previous works~\cite{mildenhall2021nerf,lu2023urban,barron2022mip}, we evaluate the novel view synthesis performance based on three commonly-used metrics, \ie, peak signal-to-noise ratio (PSNR), structural similarity index measure (SSIM)~\cite{SSIM}, and the learned perceptual image patch similarity (LPIPS)~\cite{LPIPS}.

\subsection{Comparisons with the State-of-the-art Methods}
\label{sec:comp}

\begin{table}[!t]
        \caption{\label{tab:kitti_q} Quantitative results on the KITTI-360.}
\begin{center}
	\begin{tabular}{lcccccc}  
	\toprule
	Method & Input  & PSNR $\uparrow$ & SSIM $\uparrow$  & LPIPS $\downarrow$    \\  
 \midrule
	  NeRF-W~\cite{martin2021nerf} & I & 22.77 & 79.40 & 44.60	\\
        NGP~\cite{muller2022instant}+L~\cite{rematas2022urban} & I+L & 23.36 & 85.05 & 35.63	\\
        TensoRF~\cite{chen2022tensorf} & I & 23.55 & \underline{85.14} & 36.37	\\
        K-Planes~\cite{fridovich2023k} & I & \underline{23.63} & 84.94 & 37.00	\\
	Point-NeRF~\cite{xu2022point} & I+L & 21.54 & 79.30 & 40.60	\\ 
        Tetra-NeRF~\cite{kulhanek2023tetra} & I+L & 21.73 & 80.22 & 43.60	\\ 
        Mip-NeRF 360~\cite{barron2022mip} & I & 23.27 & 83.60 & 35.50	\\ 
        DNMP~\cite{lu2023urban} & I+L & 23.41 & 84.60 & \textbf{30.50}	\\ 
        \midrule
	Ours & I+L & \textbf{23.71} & \textbf{85.21} & \underline{34.27} \\
	\bottomrule
	\end{tabular}
	\end{center}
\end{table}

\begin{table}[!t]
        \caption{\label{tab:argo_q} Quantitative results on the Argoverse2.}
	\begin{center}
	\begin{tabular}{lcccccc}  
	\toprule
	Method & Input  & PSNR $\uparrow$ & SSIM $\uparrow$  & LPIPS $\downarrow$    \\  
 \midrule
	NGP~\cite{muller2022instant} & I & 28.40 & 83.74 & 40.71	\\
	  NGP~\cite{muller2022instant}+L~\cite{rematas2022urban} & I+L & 28.64 & 84.07 & \underline{40.37}	\\
	TensoRF~\cite{chen2022tensorf} & I & \underline{30.05} & 84.09 & 42.31 \\
	K-Planes~\cite{fridovich2023k} & I & 29.26 & 83.56 & 41.86	\\
	Tetra-NeRF~\cite{kulhanek2023tetra} & I+L & 29.59 & \underline{84.33} & 42.68	\\ \midrule
	Ours & I+L & \textbf{30.62} & \textbf{85.59} & \textbf{38.32} \\
	\bottomrule
	\end{tabular}
	\end{center}
\end{table}

\begin{figure*}[!t]
    \centering
    \begin{minipage}{0.19\textwidth}
        \includegraphics[width=\linewidth]{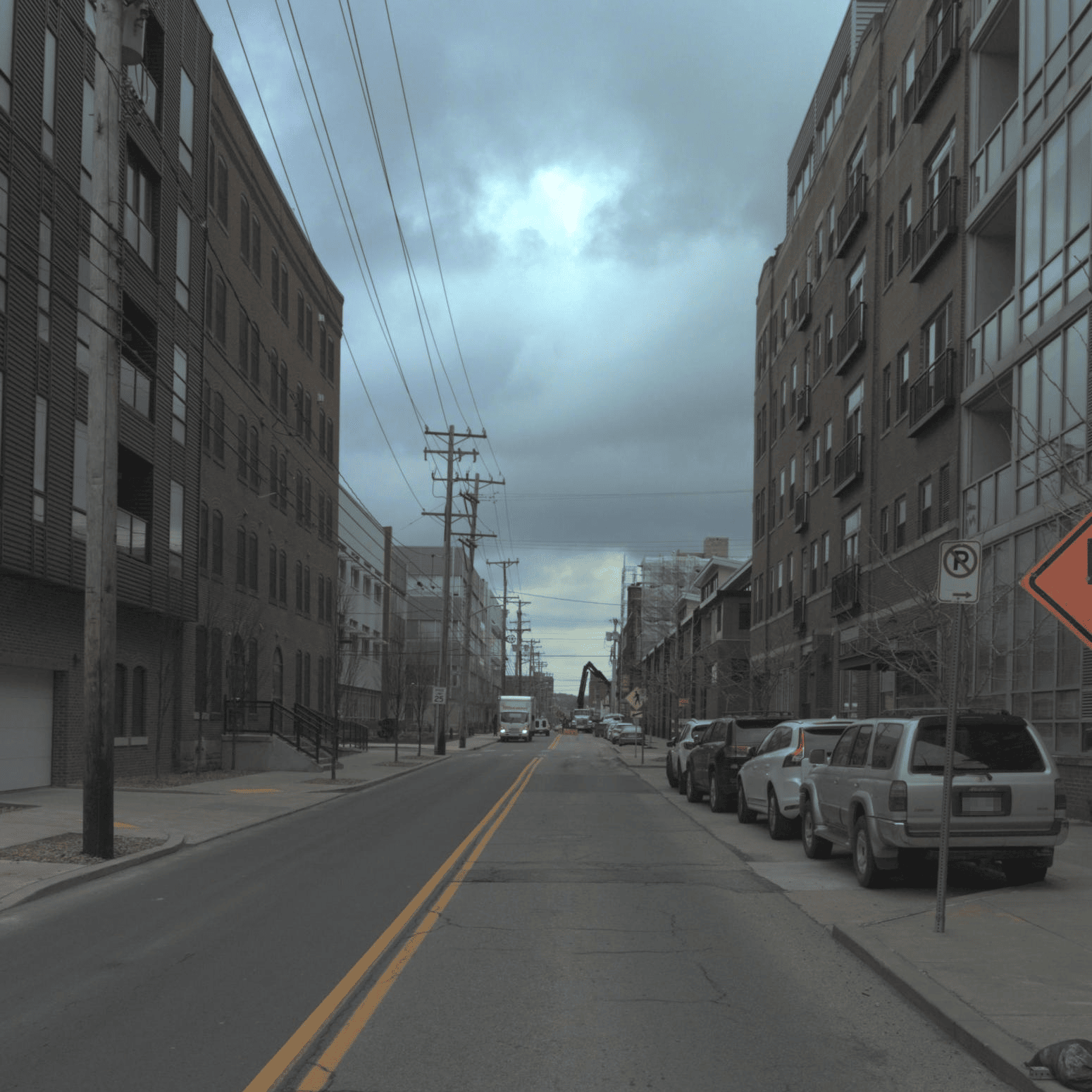}
    \end{minipage}
    \begin{minipage}{0.19\textwidth}
        \includegraphics[width=\linewidth]{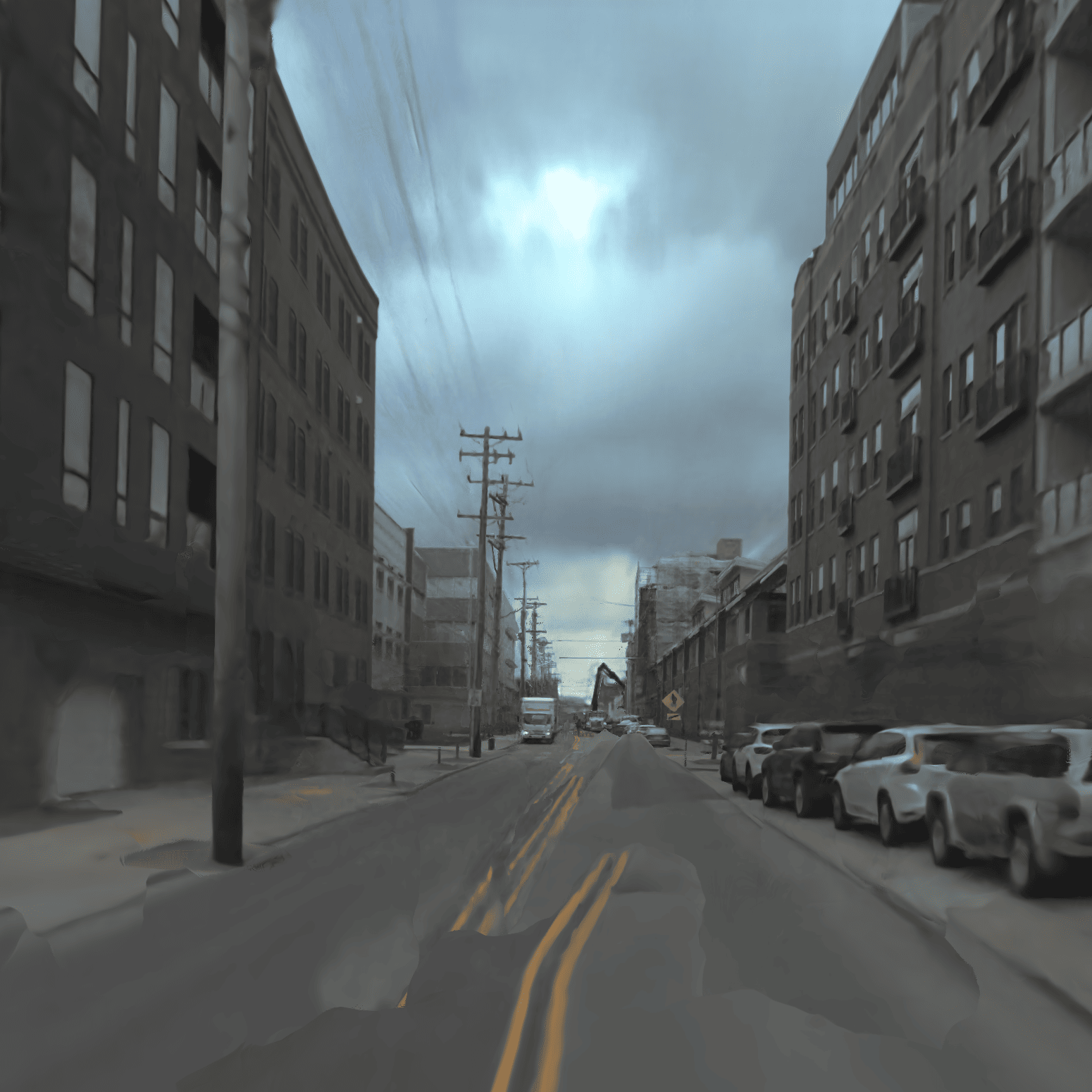}
    \end{minipage}
    \begin{minipage}{0.19\textwidth}
        \includegraphics[width=\linewidth]{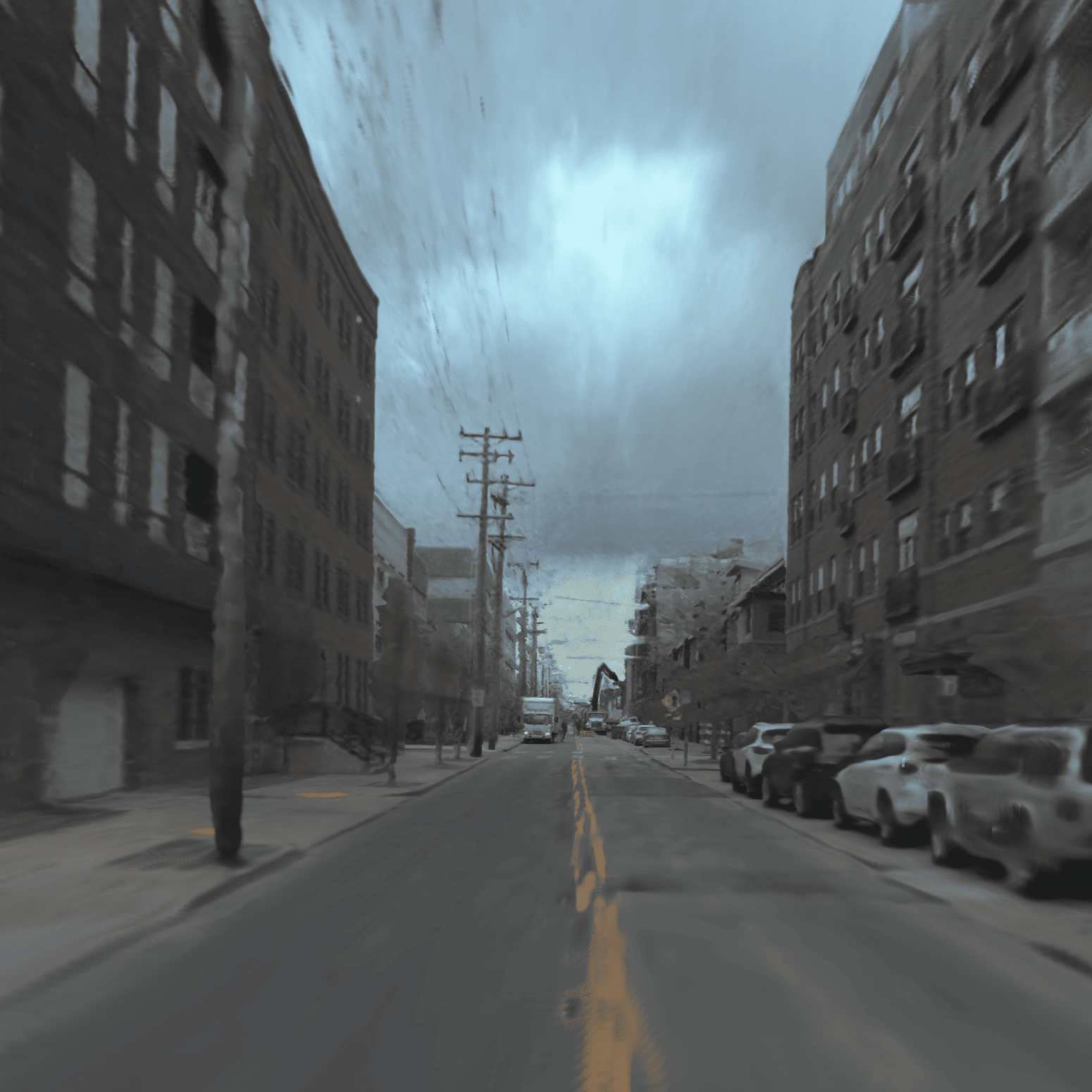}
    \end{minipage}
    \begin{minipage}{0.19\textwidth}
        \includegraphics[width=\linewidth]{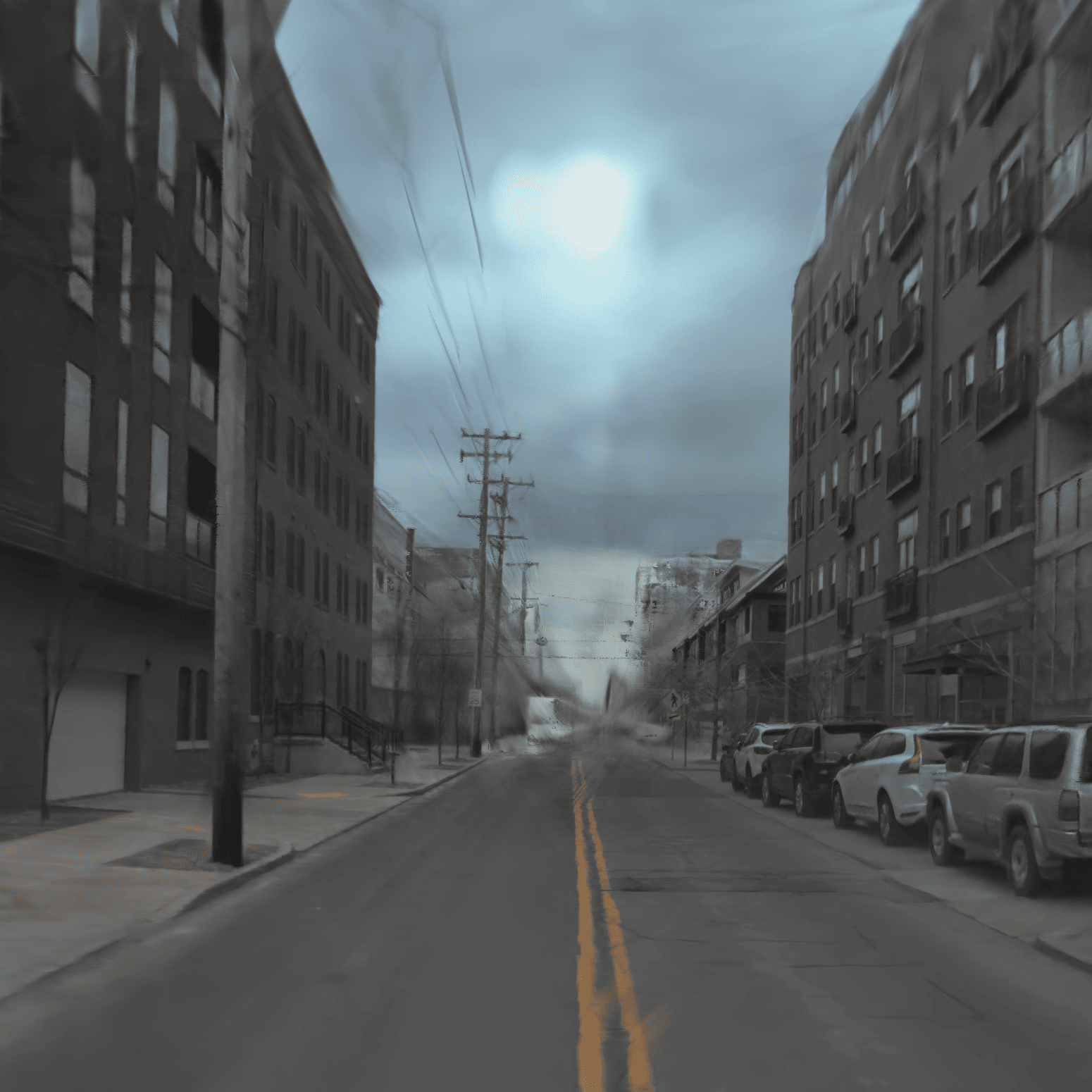}
    \end{minipage}
    \begin{minipage}{0.19\textwidth}
        \includegraphics[width=\linewidth]{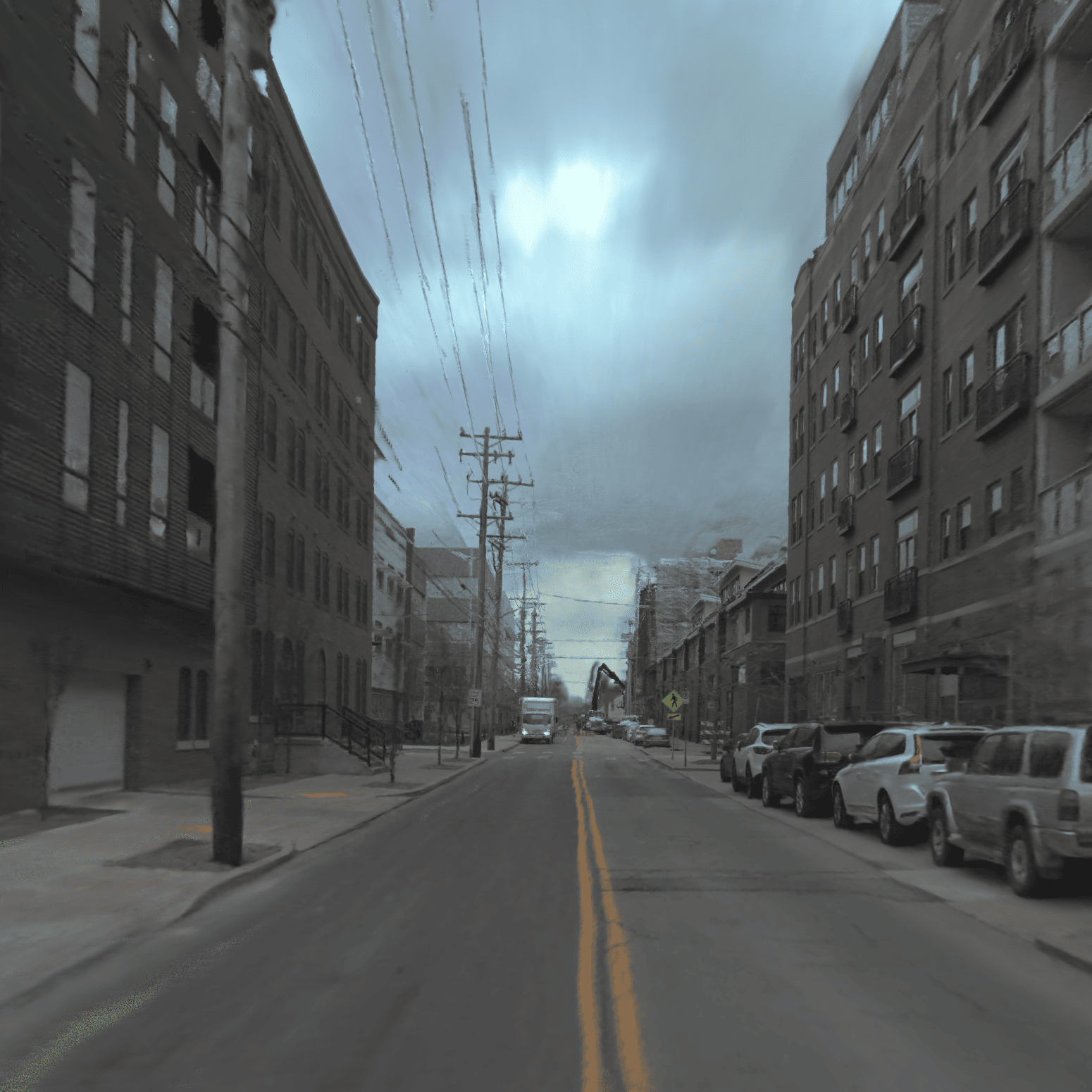}
    \end{minipage}
    
    \vspace{1mm}

    \begin{minipage}{0.19\linewidth}
        \centering
        Reference View
    \end{minipage}
    \begin{minipage}{0.19\linewidth}
        \centering
        K-Planes~\cite{fridovich2023k}
    \end{minipage}
    \begin{minipage}{0.19\linewidth}
        \centering
        NGP~\cite{muller2022instant}+L~\cite{rematas2022urban}
    \end{minipage}
    \begin{minipage}{0.19\linewidth}
        \centering
        Tetra-NeRF~\cite{kulhanek2023tetra}
    \end{minipage}
    \begin{minipage}{0.19\linewidth}
        \centering
        Ours
    \end{minipage}
    \vspace{-2mm}
    \caption{\label{fig:extra}Extrapolation results on the Argoverse2. Based on the reference view, we moved the camera 2 meters to the left.}
\end{figure*}

\noindent{\bf Quality of Novel View Synthesis.}
To analyze the rendering quality of our model, we present the novel view synthesis metrics on KITTI-360 and Argoverse2 in Tab.~\ref{tab:kitti_q} and Tab.~\ref{tab:argo_q}. 
Due to the truncation of height in KITTI-360's point clouds, there's a significant portion of missing coverage. This results in many foreground areas remaining uninitialized after LiDAR initialization. Despite the challenge, we still surpass the state-of-the-art methods in terms of PSNR and SSIM.
It should be noted that training DNMP~\cite{lu2023urban} is far less efficient than ours as it requires a mesh optimization stage before training the NeRF. In contrast to the KITTI-360, the point cloud coverage of Argoverse2 is much more comprehensive. Consequently, our method exhibits clear superiority across all metrics. 
We also display the rendered images and depth maps in Fig.~\ref{fig:dep} for a qualitative comparison. Our depth maps offer richer details with more consistent transitions, effectively capitalizing on the prior information provided by the point clouds.
Furthermore, in Fig.~\ref{fig:extra}, we showcase the extrapolation results of our method. Extrapolation is essential for closed-loop simulation systems. When SDVs exhibit behaviors different from the recorded data, the simulator should be capable of generating observations from corresponding views. Thanks to the refined depth and color decomposition achieved by our method, our extrapolation results are considerably superior compared to other techniques.

\begin{table}[!t]
        \caption{\label{tab:argo_s}Comparison of training and rendering speed with other methods. In the table, ``TT'' refers to the training time, and ``R-FPS'' refers to the rendering fps.}
        \vspace{-2mm}
	\centering
		\scalebox{0.85}{
		\begin{tabular}{l | c c | c c| c c}
		 \toprule
		 \multirow{2}{*}{Method} & \multicolumn{2}{c|}{133e2e0b (156m)} & \multicolumn{2}{c|}{2aea7bd1 (139m)}  & \multicolumn{2}{c}{b1a98ad6 (106m)} \\
		  & TT(m) & R-FPS & TT(m) & R-FPS  & TT(m) & R-FPS  \\
		 \midrule
		NGP~\cite{muller2022instant} & 2.62 & 0.19 & 4.32 & 0.19 & 3.61 & 0.17	\\
    	  NGP~\cite{muller2022instant}+L~\cite{rematas2022urban} & 2.27 & 0.22 & 3.88 & 0.23 & 3.17 & 0.19	\\
    	TensoRF~\cite{chen2022tensorf} & 7.65 & 0.15 & 7.10 & 0.15 & 8.64 & 0.15 \\
    	K-Planes~\cite{fridovich2023k} & 13.13 & 0.23 & 12.92 & 0.23 & 10.74 & 0.23	\\
    	Tetra-NeRF~\cite{kulhanek2023tetra} & 44.64 & 0.02 & 129.98 & 0.02 & 60.01 & 0.02	\\ \midrule
    	Ours & \textbf{0.76} & \textbf{2.30} & \textbf{1.07} & \textbf{2.56} & \textbf{0.88} & \textbf{2.18} \\
		
		 \bottomrule
		\end{tabular}
		}
\vspace{-2mm}
\end{table}

\noindent{\bf Training \& Rendering Cost.}
Tab.~\ref{tab:argo_s} showcases the advantages of our proposed method in terms of training and rendering speed. 
For a fair comparison of training speed, we report the training time for each method to converge to the same PSNR as NGP~\cite{muller2022instant}. Specifically, we choose 3 sequences from Argoverse2~\cite{wilson2023argoverse} in this section for comparisons. For sequences 133e2e0b, 2aea7bd1, and b1a98ad6, we use PSNR values of 28.5, 27.0, and 28.0, respectively, to report the training time.
We achieve significant speed improvements in all test scenes. As can be seen from the table, the combination of our proposed LiDAR Initialization (LI) and Hybrid Scene Representation (HSR) yields more than 4 times boost in terms of convergence speed of our method compared with NGP~\cite{muller2022instant}.
Additionally, because we model the occupancy information explicitly, our instructive sampling reduces the number of sample points during rendering and saves time on MLP inference, thereby achieving a rendering speed that is 10 times faster. We will discuss in detail the impact of each component on speed in the following Sec.~\ref{sec:ablation}. We also show the results in the private dataset in Tab.~\ref{tab:ts_s} to demonstrate the advantage of training speed on large-scale scenes. For the 600m scene, the proposed method trained within 1 minute exceeds the performance of NGP+L trained for 5 minutes. These results convincingly verify the efficiency of our proposed method.

\begin{table}[!t]
        \caption{\label{tab:ts_s} Results on the private dataset. We presented metrics for 1-minute and 5-minute training.}
        \vspace{-2mm}
	\centering
		\scalebox{1}{
		\begin{tabular}{l | c c | c c}
		 \toprule
		 & \multicolumn{2}{c|}{1min} & \multicolumn{2}{c}{5min}    \\
		 v1 (600m) & PSNR $\uparrow$ & LPIPS $\downarrow$ & PSNR $\uparrow$ & LPIPS $\downarrow$  \\
		 \midrule
    	  NGP~\cite{muller2022instant}+L~\cite{rematas2022urban} & 27.95 & 41.68 & 29.03 & 38.43	\\
    	 Ours & 30.48 & 37.14 & 31.69 & 33.14 \\
		 \midrule
		 v2 (1000m) & PSNR $\uparrow$ & LPIPS $\downarrow$ & PSNR $\uparrow$ & LPIPS $\downarrow$  \\
		 \midrule
    	  NGP~\cite{muller2022instant}+L~\cite{rematas2022urban} & 27.34 & 45.06 & 29.27 & 41.33	\\
    	 Ours & 28.79 & 40.65 & 30.04 & 36.83 \\
		 \bottomrule
        \end{tabular}
        }
        \vspace{-2mm}
\end{table}

\begin{table*}[!t]
        \caption{\label{tab:abla}Ablation studies for proposed components on the Argoverse2.}
	\centering
		\scalebox{1}{
  \setlength{\tabcolsep}{1.8mm}{
		\begin{tabular}{l | c c c c | c c c c| c c c c}
		 \toprule
		\multirow{2}{*}{Variant} & \multicolumn{4}{c|}{133e2e0b} & \multicolumn{4}{c|}{2aea7bd1}  & \multicolumn{4}{c}{b1a98ad6} \\
		 & PSNR & \#Sample/Ray & TT(m) & R-FPS   & PSNR & \#Sample/Ray & TT(m) & R-FPS & PSNR & \#Sample/Ray & TT(m) & R-FPS  \\
		 \midrule
		w/o HSR & 29.23 & 77.12 & 3.93 & 0.34 & 26.36 & 51.73 & 8.71 & 0.56 & 28.89 & 64.01 & 4.56 & 0.40	\\
    	  w/o LI & 25.65 & 39.42 & Inf. & 0.72 & 25.86 & 34.48 & Inf. & 0.76 & 24.61 & 47.19 & Inf. & 0.60	\\
          w/o CD & \textbf{31.86} & 27.07 & \textbf{0.57} & 1.29 & \textbf{30.42} & 21.40 & \textbf{0.70} & 1.53 & 31.42 & 24.06 & \textbf{0.82} & 1.27	\\ 
    	Ours & 31.74 & \textbf{12.06} & 0.76 & {\bf 2.30} & 30.04 & \textbf{11.26} & 1.07 & {\bf 2.56} & \textbf{31.55} & \textbf{12.61} & 0.88 & {\bf 2.18} \\
		
		 \bottomrule
		\end{tabular}
		}}
\end{table*}

\begin{table}[!t]
\centering
\caption{\label{tab:abl_bgm}Ablation studies for background modeling in terms of PSNR.}
    \begin{tabular}{l|c c c}
    \toprule
        Background Modeling &  133e2e0b &  2aea7bd1 & b1a98ad6 \\
    \midrule
        Enlarged Scene Box~\cite{muller2022instant} & 31.43& 24.46&30.02\\
        Sphere Background~\cite{rematas2022urban} &30.94 & 26.34&29.21 \\
        Ours & {\bf 31.74} & {\bf 30.04} & {\bf 31.55} \\
    \bottomrule
    \end{tabular}
\end{table}

\subsection{Ablation Study}
\label{sec:ablation}
\noindent{\bf Analysis on Proposed Components.} To analyze the roles of different components, we present the novel view synthesis quality and efficiency in Tab~\ref{tab:abla}. The PSNR values in the table are obtained under the standard training setting on test sets, while the training time reflects how long the models converge to the same PSNR as NGP~\cite{muller2022instant}, same as in Tab.~\ref{tab:argo_s}. Based on our method, we respectively remove corresponding modules to analyze the differences in outcomes, including Hybrid Scene Representation (HSR), LiDAR Initialization (LI), and Color Decomposition (CD).
From the table, it's evident that the most significant impacts on training and rendering speed come from HSR and LI. Specifically, 
when HSR is ablated, the model uses NGP's multi-scale hash grids to store features and uses separate MLPs to implicitly model both volume density and color. In large-scale scenes, high-resolution hash grids cause severe hash collisions. NGP typically relies on MLPs to handle these collisions, making the spatial representation ambiguous and leading to declines in novel view synthesis performance. Additionally, the density MLP introduces prediction overheads, resulting in a significant drop in training and rendering speeds.
On the other hand, without LI, plus the absence of an MLP in our model to fit density, it cannot converge to NGP's PSNR. This results in the training time being shown as ``Inf.''. Furthermore, due to the absence of initialization and fitting capabilities, the model struggles to represent space sparsely, leading to a substantial number of samples per ray. This, in turn, leads to a noticeable decline in both rendering speed and quality. Finally, CD enables the model to focus its representation on most non-translucent and non-reflective areas without requiring excessive sample points for description. However, unsupervised decomposition slightly increases the fitting complexity, leading to a minor drop in PSNR. This represents a trade-off between performance and efficiency.

\noindent{\bf Analysis on Background Modeling.} Here, we compare our background modeling strategy with the aforementioned proposals, \ie, ``Enlarged Scene Box'' used by~\cite{muller2022instant} and ``Shpere Background'' adopted by~\cite{rematas2022urban}. The comparative results are shown in Tab.~\ref{tab:abl_bgm}. From the table, it is observed that our background modeling approach consistently outperforms the others by a clear margin, which verifies the advantage of our proposed strategy in autonomous driving scenarios.

\begin{table}[!t]
\centering
\caption{\label{tab:abl_cm}Ablation studies for color modeling.}
\setlength{\tabcolsep}{1.2mm}
    \begin{tabular}{l | c c | c c| c c}
		 \toprule
		 \multirow{2}{*}{Color Modeling} & \multicolumn{2}{c|}{133e2e0b} & \multicolumn{2}{c|}{2aea7bd1}  & \multicolumn{2}{c}{b1a98ad6} \\
		  & PSNR & R-FPS & PSNR & R-FPS  & PSNR & R-FPS  \\
    \midrule
        SH (degree=0) & 27.42 &\textbf{3.83}  &22.80 &\textbf{3.91}  & 28.25& \textbf{3.64} \\
        SH (degree=1) & 29.06 & 3.33 &24.33 & 3.43 & 28.98& 2.96 \\
        SH (degree=2) & 29.14 & 2.64 &24.68 & 2.79 &29.29 &2.43 \\
        Ours & {\bf 31.74} & 2.30 & {\bf 30.04} & 2.56& {\bf 31.55} & 2.18\\
    \bottomrule
    \end{tabular}
\end{table}

\noindent{\bf Analysis on Color Modeling.} Some recent work~\cite{yu2021plenoctrees,fridovich2022plenoxels} propose to use spherical harmonic (SH) to model scene appearance explicitly for efficiency, avoiding the need for a color MLP. To compare the effectiveness of our proposed color modeling with SH representations, we conduct ablation studies and show the results in Tab.~\ref{tab:abl_cm}. It is observed that SH representations achieve fast rendering speed thanks to explicit color modeling. However, due to the complexity inherited from large-scale scenarios, SH with limited grid resolution cannot achieve fine rendering quality compared with our hybrid representations. In summary, our method enjoys efficient training and rendering via explicit modeling of scene geometry while maintaining rich expression ability for scene appearance via implicit color modeling.

\section{CONCLUSION}

This paper introduces \Name, an efficient novel view synthesis framework for outdoor scenes that integrates point clouds and images. The proposed method utilizes point clouds for a swift initialization of a sparse representation of the scene, achieving significant performance and speed enhancements. By modeling the background more effectively, we reduce the representational strain on the foreground. Finally, through color decomposition, we model view-dependent and view-independent colors separately, enhancing the model's extrapolation capability. Extensive experiments on various autonomous driving datasets demonstrate that our method outperforms the previous state-of-the-art techniques in both performance and efficiency.

{\flushleft \bf Acknowledgments.} This work was supported in part by NSFC (62322113, 62376156), Shanghai Municipal Science and Technology Major Project (2021SHZDZX0102), and the Fundamental Research Funds for the Central Universities.





\bibliographystyle{IEEEtran}
\bibliography{IEEEabrv,bibfile}

\end{document}